\documentclass{article}

    \usepackage[preprint]{neurips_2020}

\usepackage[utf8]{inputenc} % allow utf-8 input
\usepackage[T1]{fontenc}    % use 8-bit T1 fonts
\usepackage{hyperref}       % hyperlinks
\usepackage{url}            % simple URL typesetting
\usepackage{booktabs}       % professional-quality tables
\usepackage{amsfonts}       % blackboard math symbols
\usepackage{nicefrac}       % compact symbols for 1/2, etc.
\usepackage{microtype}      % microtypography
\usepackage{graphicx}
\usepackage{listings}

\usepackage{amsmath}
\usepackage{appendix}
\usepackage{float}

\title{SPINE: Soft Piecewise Interpretable Neural Equations}

\author{Jasdeep Singh Grover \\
  Department of Computer Engineering\\
  KJ Somaiya College of Engineering\\
  Vidyavihar, Mumbai -400077 \\
  \texttt{jasdeepsingh.g@somaiya.edu} \\
  \And
  Harsh Minesh Domadia \\
  Department of Computer Engineering\\
  KJ Somaiya College of Engineering\\
  Vidyavihar, Mumbai -400077 \\
  \texttt{harsh.domadia@somaiya.edu} \\
  \AND
  Raj Anant Tapase \\
  Department of Computer Science and Engineering\\
  National Institute of Technology Karnataka, Surathkal\\
  Mangalore- 575025 \\
  \texttt{raj.tapase1@gmail.com} \\
  \And
  Grishma Sharma\\
  Department of Computer Engineering\\
  KJ Somaiya College of Engineering\\
  Vidyavihar, Mumbai -400077 \\
  \texttt{grishma.sharma@somaiya.edu} \\
  }

\begin{document}

\maketitle

\begin{abstract}

Relu Fully Connected Networks are ubiquitous but uninterpretable because they fit piecewise linear functions emerging from multi-layered structures and complex interactions of model weights. This paper takes a novel approach to piecewise fits by using set operations on individual “pieces”(parts). This is done by approximating canonical normal forms and using the resultant as a model. This gives special advantages like (a)strong correspondence of parameters to pieces of the fit function(High Interpretability); (b)ability to fit any combination of continuous functions as pieces of the piecewise function(Ease of Design); (c)ability to add new non-linearities in a targeted region of the domain(Targeted Learning); (d)simplicity of an equation which avoids layering. It can also be expressed in the general max-min representation of piecewise linear functions which gives theoretical ease and credibility.  This architecture is tested on simulated regression and classification tasks and benchmark datasets including UCI datasets, MNIST, FMNIST and CIFAR 10. This performance is on par with fully connected architectures. It can find a variety of applications where fully connected layers must be replaced by interpretable layers.

\end{abstract}

\section{Introduction}

In many architectures, fully connected layers make the final predictions after feature extraction. This is seen in Resnet[1] in computer vision, ULMfit[2] in Natural Language Processing and many reinforcement learning problems like disease prognosis(symptom checking) as shown in [3]. Despite their ubiquity, understanding them and the functions they fit is challenging. Some attempts include: (a)gradient-based methods like Smooth Gradients[4] and Grad CAM[5]; (b)perturbation based methods like LIME[6]; (c)methods based on hidden representations like Inverted Representations[7]; and, (d)attempts to understand the loss function and training process like [8]. Comparatively, lesser work is done in understanding the fit function, like in [9]. Such an analysis helps in better understanding of the architecture, its properties and the fit function. Relu Fully Connected Neural Networks(FCNNs) are essentially piecewise linear(every part is linear) functions which behave as Universal Function Approximators as stated in [10] and [11]. This is an important observation analysed very well in [9].

Another important direction of work involves building of models which are self interpretable and can explain the decisions which they make. One such example is Neural Backed Decision Trees[23] where the final classifications are explained using hierarchical features and hierarchy in classes. Simultaneously, new models with easy to understand structure and decision boundaries are also built. Such examples include Deep Neural Decision Trees[24], Disjunctive Normal Networks[25] and Gated Linear Networks[26]. Works involving Neural Decision Trees generally try to explain their decisions using soft decisions at every node using sigmoid function. In various cases, interpretable nodes are generated which provide interpretability to the architecture as a whole. Thus models can be interpreted in terms of their decision boundaries and weights providing novel benefits in terms of interpretability and even online learning as shown in Gated Linear Networks.

In another recent research [12], it is shown that given certain inequations and set operations on them, a resultant inequation can be obtained which represents the region of the domain as dictated by the set operations. This is done by approximating every set operation with a differentiable operation. The approximation accuracy is regulated by a parameter $a$. Higher its value, closer is the approximation. This result provides a new means to generate interpretable models where weights and decision boundaries can be interpreted and controlled individually.

Considering the complexities of FCNNs, this paper takes a different approach to piecewise fits. Piecewise fits are represented as a combination of set operations (union and intersection) approximated by operations in [12]. Every “piece” of the piecewise function is considered as an inequality or a set. This makes the piecewise function a combination of set operations which can be represented in the disjunctive normal form. This is very different from FCNNs and activation functions as neurons do not feed into other neurons, rather organise themselves according to the set operations. This gives special advantages like: (a) parameters of every “piece” affect the corresponding “piece” and not other “pieces”, thus they can be easily understood, (b) “pieces” need not be linear and can be any continuous function(even sinusoidal) or a combination of continuous functions(like linear and quadratic), (c) if the model does not fit well, then more sets can be added where needed thus improving the fit; and, (d) all this can be expressed with one linear layer and simple mathematical operations. The model can further be simplified to the general max-min form of piecewise linear functions for linear pieces. This simplicity may give faster speeds and ease of designing and debugging. This paper shows how this architecture performs regression and classification instead of FCNNs with identical results as well.

\section{Mathematical formulation}

FCNNs fit piecewise linear functions when they use Relu and LeakyRelu activation functions. This inspired the use of set operations to do the same as described in this section.
 %%%%%%%%%%%%%%%%% Figure 1 A section created using an intersection of 3 line segments && Union of 3 intersected regions to estimate sin(x)%%%%%%%%%%%
\begin{figure}[ht]
  \centering
  \includegraphics[width=3.5in,height=0.5in]{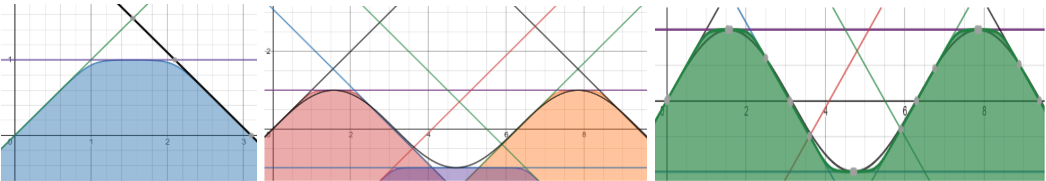}
%   \fbox{\rule[-.5cm]{0cm}{4cm} \rule[-.5cm]{4cm}{0cm}}
  \caption{\protect\raggedright Mathematical Formulations: Left(a): Intersection of 3 lines generating a polytope. It is expressed as the following Log-exp approximation: \( y<=ln \left( 1/ \left( e^{-10 \left( x+0 \right) }+e^{-10 \left( 0x+1 \right) }+e^{-10 \left( -x+3.1 \right) } \right)  \right) /10 \); Middle(b): Multiple Polytopes used; Right(c): Union of 3 polytopes to estimate sin(x). Log-exp approximation of union of polytopes:
  $y<=ln(1/(e\textsuperscript{-10(x+0)}+e\textsuperscript{-10(0x+1)}+e\textsuperscript{-10(-x+3.1)})+1/(e\textsuperscript{-10(x-4.1)}+e\textsuperscript{-10(0x+1)}+e\textsuperscript{-10(-x+5.5)})+(e\textsuperscript{-10(x-6.3)}+e\textsuperscript{-10(0x-1)}+e\textsuperscript{-10(-x+9.4)}))/10$; Try these in 2D![13], 3D![14];}
\end{figure}
%%%%%%%%%%%%% Figure 1 ends %%%%%%%%%%%%%%%%%%%%%%%%%%%%%
%FCNNs fit piecewise linear functions when they use Relu and LeakyRelu activation functions. All piecewise linear functions can be expressed as:
Linear inequations can intersect to give convex piecewise linear functions(generally polytopes) as shown in Figure 1(a). Union of these polytopes can then give the expected piecewise linear function as shown in Figure 1(b, c). The intersection operation is equivalent to pointwise min() and union to pointwise max() of intersected regions. This gives the max-min form for piecewise linear functions as shown in Eq. 1.
\begin{equation}max(min( w\textsuperscript{t}\textsubscript{ij}x+b\textsubscript{ij}, \exists i \forall j), \forall i)\end{equation} 
In the max-min form, $w (w \in R^n)$, the t in superscript indicates transpose and $b (b \in R)$ are learnable weights and biases, and $x (x \in R^n)$ is the input. The output $w\textsuperscript{t}\textsubscript{ij}x+b\textsubscript{ij}$ is split into non-null sets represented by $i$ and every element of these sets is represented by $j$. Min() is taken over every set followed by max() over all min outputs. The max-min form is a piecewise linear function and $w\textsubscript{ij}\textsuperscript{t}x+b\textsubscript{ij}$(component functions) are “pieces” of the output. It is also independently derived and detailed in [15].

Empirically, the max-min form does not train well probably due to 0 gradients for unexpressed components as shown in Appendix E. To handle this, all unions and intersections are expressed in the disjunctive normal form(conjunctive can also be used). Approximations provided in [12] are then applied to give Eq. 2 which is called the log-exp form (as derived in Appendix A). Log-exp form is an approximate piecewise function with $f\textsubscript{ij}(x,\theta)$ as pieces just like linear functions in max-min form. Here, $a$ is a crucial hyperparameter obtained from [12], $m$ is the number of intersections and summation over $j$ represents intersection operation, $n$ is the number of unions and summation over $i$ represents the union operation, and $\theta$s are learnable parameters. Here, $a$ generally ranges from 0.1 to 20 and controls the accuracy of approximation of unions and intersections, thus higher its value, closer is the log-exp form to the max-min form and vice versa. When $a$ tends to infinite the approximation becomes exact (as shown in Appendix B). The equations in Figure 1 show how linear components are intersected and how union is taken over the generated polytopes. Another set of works[16][17] use the log-sum-exp(softmax) form for creating neural networks for regression tasks. This is a special case of the log-exp form when only unions(max operations) are used. Work [16] also provides theoretical derivations for error bounds and approximation behaviours which can be extended to this paper in future.

\begin{equation}y=ln \left(  \sum _{i=1}^{n}1/ \left(  \sum _{j=1}^{m}e^{-af\textsubscript{i,j} \left( x,  \theta  \right) } \right)  \right) /a \end{equation}

The same idea is extended to higher dimensions. Polytopes in higher dimensions need many more linear components, but very good results can be obtained even with lesser components. This is tested empirically and depends on the nature of the dataset. Continuous non-linear component functions(even sinusoidal) can also be used in the log-exp form as it is fundamentally based on set operations. Parameters have much more effect on their corresponding components and much lesser on other components in the log-exp form. $a$ controls the effect of parameters across different components, higher its value lower is the cross-component effect. Particular components being used for each input can also be identified using gradients, perturbations or by simply applying the max-min form instead of log-exp for large $a$. Thus, the use of set operations instead of nested operations allows simplicity of formulation, interpretability of parameters, identification of fit components for every input and ease in designing(controlling the structure of the fit function) and debugging(correcting the structure of the  fit function as discussed in Targeted Learning Section).

The main focus of this paper is on performing regression and classification using the log-exp form but the ideas of Disjunctive Normal Networks and Neural Decision Trees can also be implemented using this mathematical formulation with high accuracy. This is demonstrated in the Appendix L.

\section{Architecture}
%%%%%%%%%%%%%%%%%%%%%%% Figure 2 Use as a Regressor and a Classifier
%%%%%%%%%%%%%%%%%%%%%%
\begin{figure}[ht]
  \centering
  \includegraphics[width=5in]{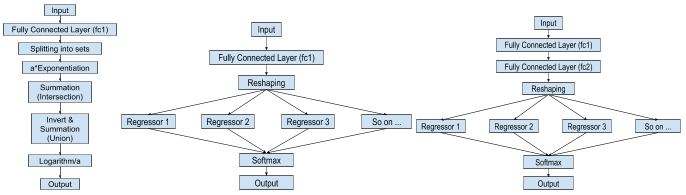}
%   \fbox{\rule[-.5cm]{0cm}{4cm} \rule[-.5cm]{4cm}{0cm}}
  \caption{Use as a Regressor and a Classifier. Left(a): Piecewise Linear log-exp Regressor; Middle(b): Classifier; Right(c): Modified classifier for large inputs. The regressors in classifier architectures have parts coming after linear layers.}
%   \vspace{-5mm}
\end{figure}
%%%%%%%%%%%%%%%% Figure 2 ends here %%%%%%%%%%%%%%%%%%%%%%%%
\subsection{Use as a Regressor}
The log-exp form can be used to learn a target function by minimizing the mean squared error using gradient descent or other optimizers and thus learning $\theta$s. A piecewise linear function is fit when $f\textsubscript{ij}(x,\theta)$ are linear, just like Relu FCNNs. The complete architecture is available in Figure 2(a). The advantages of using set operations here are: (a) continuous component functions can be selected(like linear, sinusoidal etc. depending on data) or even combinations can be taken(like linear and quadratic, as shown in Appendix G); (b) model parameters strongly correspond to respective components, making the fit interpretable; and, (c) parameters and components used for prediction of specific examples can be identified and manipulated. These advantages are shown in upcoming sections. Note that the log-exp form is the architecture itself and not just an activation function.

\subsection{Use as a Classifier}

FCNNs share features among all classes until the final layer, hence reduce the model size and allow weight sharing between classes, which is not used in this paper. Every class here has its own regressor which is akin to the final layer neuron of that class in FCNNs. Softmax is then used on the regressor outputs to make predictions as shown in Figure 2(b) and various loss functions can then be used for training. This increases parallelism as there is no concept of layers thus reducing the number of serial matrix multiplications. It also enhances interpretability as each class has its own regressor. To avoid fast growth in parameters with classes, regressors can also share polytopes (shown in Appendix H).

For implementation, fully connected layers in various regressors are merged into a single layer followed by reshaping to separate the outputs of every regressor. Parameters can be reduced by applying a linear layer without activation function and common to all regressors, to the input as shown in Figure 2(c). This layer can later be merged with the first layer as there is no activation function, thus the properties of the architecture can be retained. This was done for CIFAR 10 dataset.

\section{Experimentation and Analysis:}

\subsection{As a Regressor}

The regressor architecture is tested on a simulated dataset and the output of an FCNN trained on the MNIST dataset. The details and results are provided in the following subsections:

\subsubsection{On Simulated Dataset:}

A simulated dataset is generated as shown in Figure 4. Two log-exp variants using linear and sinusoidal components are trained and compared with FCNNs. Target is z-score normalised for FCNNs and linearly scaled between 0 and 1 for log-exp regressors as the two architectures work best with the corresponding preprocessing (MSEs adjusted accordingly). Input is z-score normalised for all models.

The first variant, Linear log-exp regressor model can be described as:

y = (a$\cap$b$\cap$c … Intersection $i$=3 terms)$\cup$(d$\cap$e$\cap$f)$\cup$(g$\cap$h$\cap$i)$\cup$…for Unions $u$=25 terms where a, b, c…are linear functions of $x$. The second variant has a, b, c... as sinusoids(A*sin(wx+b)+B where A, B, w and b are learnable) and $u=33$ and $i=3$.

% %%%%%%%%%%%%%%%%%%%%%% table 1 Results of Simulated Regression Data %%%%%%%%%%%%
% \begin{table}[ht]
%   \caption{Results of Simulated Regression Data}
%   \label{Table 1}
%   \centering
%   \begin{tabular}{lll}
%     \cmidrule(r){1-3}
%     Model (Type, Hyperparams)    & Rescaled Mean    & Epochs to  \\
%      & Squared Error & Converge \\
%     \midrule
%     Log-exp 1 (Linear, (25,3)) & 0.004 & 2000 \\
%     Log-exp 2 (Sin, (33,3))) & 0.02 & 2000  \\
%     Fully Connected 1 & 0.0238 & 1000  \\
%     (LeakyRelu, 1 hidden, [1,73,1]) &  & \\
%     Fully Connected 2 & 0.071 & 1000  \\
%     (LeakyRelu, 3 hidden, [1,12,15,10,1]) &   & \\
%     \bottomrule
%   \end{tabular}
% \end{table}
% %%%%%%%%%%%%%%% table 1 ends %%%%%%%%%%%%%%%%%%%%%%%
%%%%%%%%%%%%%%%%%%%%%%% Figure 3 Simulated Regression Data %%%%%%%%%%%%%%
\begin{figure}[ht]
  \centering
  \includegraphics[width=4in,height=1in]{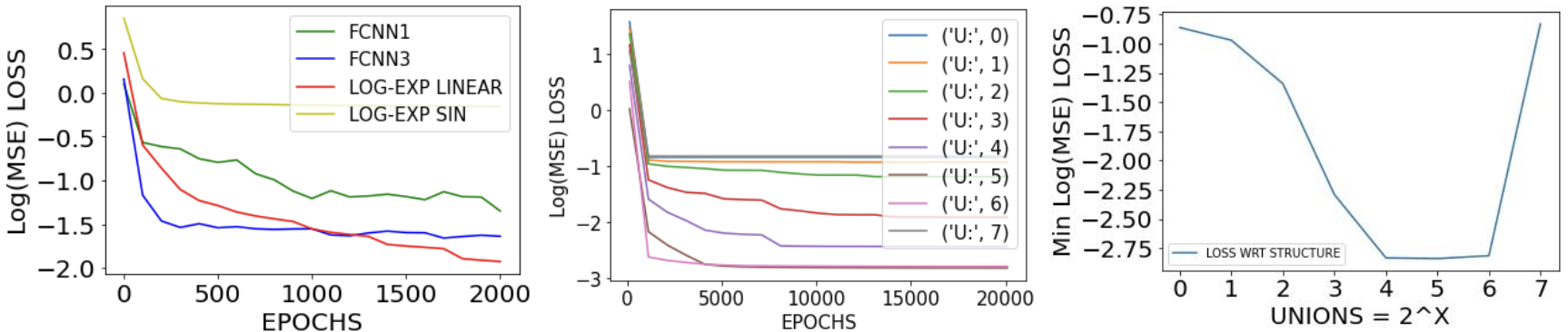}
%   \fbox{\rule[-.5cm]{0cm}{4cm} \rule[-.5cm]{4cm}{0cm}}
  \caption{Simulated Regression Convergence. Left(a): Log(MSE) vs Epochs; Middle(b): Convergence for different no. of unions u=2\textsuperscript{U} and intersections i=128/u; Right(c): Minimum loss vs no. of unions}
\end{figure}
%%%%%%%%%%%%%%%% Figure 3 ends %%%%%%%%%%%%%%%%%%%%%%%
Piecewise linear models perform better than sinusoidal ones as seen in Figure 3(a). Minimum MSEs obtained are: 0.011 for Linear log-exp, 0.29 for Sinusoidal log-exp, 0.044 for FCNN\{1,73,1\} and 0.023 for FCNN\{1,12,15,10,1\} and both FCNNs use LeakyRelu. The MSE in linear log-exp form can be reduced further till 0.0052 with 20,000 epochs. The log-exp linear models generally fit better than FCNNs having lesser losses. Number of unions and intersections is problem dependent and they are important hyperparameters as seen in Figure 3(b)(c). (More setup details are given in Appendix C and robustness to an increase in noise is analysed in Appendix F).
%%%%%%%%%%%%%%%%%%%%%%% Figure 3 Simulated Regression Data %%%%%%%%%%%%%%
\begin{figure}[ht]
  \centering
  \includegraphics[width=4in,height=1in]{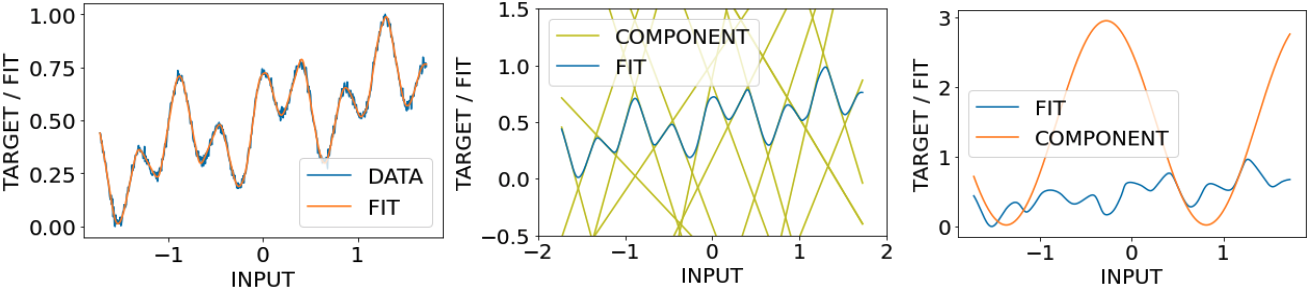}
%   \fbox{\rule[-.5cm]{0cm}{4cm} \rule[-.5cm]{4cm}{0cm}}
  \caption{Simulated Regression. Left: Fit function; Middle: Major Linear Components of the fit; Right: Major sinusoidal components(More available in Appendix J)}
\end{figure}
%%%%%%%%%%%%%%%% Figure 3 ends %%%%%%%%%%%%%%%%%%%%%%%

The function fit by the log-exp regressors can be decomposed back into corresponding components as seen in Figure 4. For linear components, slope represents weights and intercepts represent biases. Similar associations are seen in non-linear components. This makes parameters inherently interpretable. These components can be identified using perturbation analysis which perturbs parameters one by one using uniformly distributed values from 0.9 to 1.1 times their real value and output standard deviation are observed. Figure 5 shows that the effect of perturbations is extremely local in the log-exp regressors and is spread across a long part of the domain in FCNNs. This also proves that parameters significantly affect their corresponding components and not the entire fit, unlike FCNNs. The components can also be identified by using the max-min form with same parameters as log-exp form and using argmax and argmin because log-exp form approximates it (refer Appendix B).

%%%%%%%%%%%%%%%%% Figure 4 Heatmap for Perturbation analysis of presented architecture(Upper) and FCNNs(Lower) %%%%%%%%%%%
\begin{figure}[ht]
  \centering
  \includegraphics[width=5in,height=1.5in]{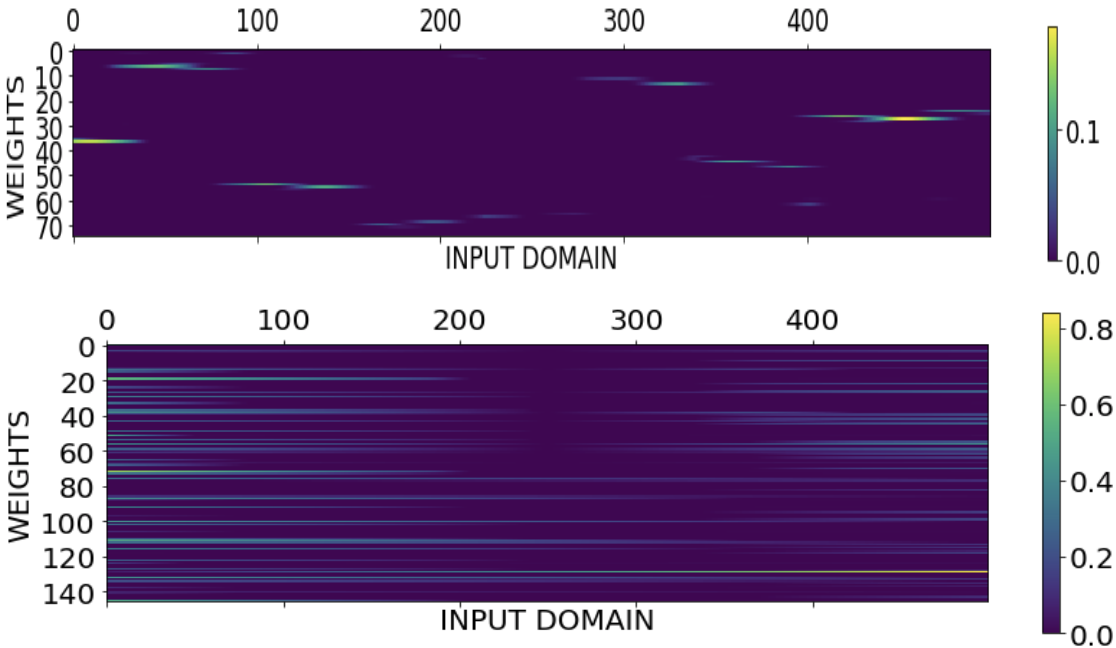}
%   \fbox{\rule[-.5cm]{0cm}{4cm} \rule[-.5cm]{4cm}{0cm}}
  \caption{Perturbation Analysis: The input space(x) is divided into 500 equal part and standard deviation of weights(y) is shown. Upper: Log-Exp form; Lower: FCNNs\{1,73,1\};}
\end{figure}
%%%%%%%%%%%%% Figure 4 ends %%%%%%%%%%%%%%%%%%%%%%%%%%%%%

\subsubsection{Mimicking a Neural Network:}

The log-exp form is also tested on functions learnt by neural nets to demonstrate that more complex high dimensional functions can be learnt as well. An FCNN with architecture [784(input), 289, 49, 10(output)] with Relu activation is trained on the MNIST dataset. The final softmax function is removed and the output neuron predicting class 0 is used as the target function. The MNIST training set is used for training and the MNIST test set for testing. The model used to approximate has $u*i$ linear component functions with $u$ union terms and $i$ intersection functions per term. 10 runs of 1500 epochs each are conducted for every pair of $u$ and $i$ and average of minimum loss in every run is taken. Figure 6 and Table 1 show results for various values of $u$ and $i$ and prove that the architecture works even for high dimensional and highly non-linear functions. The loss is generally smaller when $u$ and $i$ take similar values.

%%%%%%%%%%%%%%%%%%%Figure 5 Fitting on Neural Network %%%%%%%%%%%%%%%%%%%
\begin{figure}[ht]
  \centering
  \includegraphics[width=2.5in,height=1.2in]{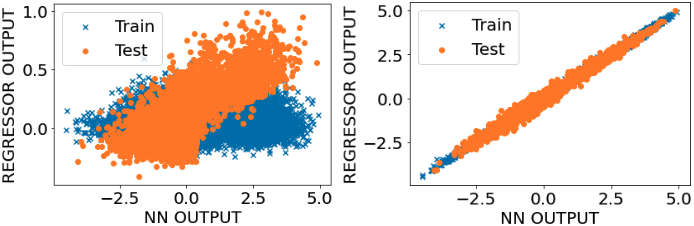}
%   \fbox{\rule[-.5cm]{0cm}{4cm} \rule[-.5cm]{4cm}{0cm}}
  \caption{Fitting on Neural Network. X-axis: Actual NN Data, Y-axis: Regressor Predictions, Blue: Training, Orange: Testing; Left: Before Training, Right: After Training}
\end{figure}
%%%%%%%%%%%%%%%%%%% Figure 5 ends here %%%%%%%%%%%%%%%%%%%%%%%%%%%%%%%%%%%

%%%%%%%%%%%%%%%%%%%% table 1 Mimicking a Fully Connected Neural Network %%%%%%%%% 
\begin{table}[ht]
  \caption{Mimicking a Fully Connected Neural Network with different u and i pairs}
%   \label{}
  \centering
  \begin{tabular}{llll}
    \cmidrule(r){1-4}
    u, (i=1024/u)    & Min. Test MSE    & u, (i=1024/u)  & Min. Test MSE\\
    \midrule
    1 & 0.0295 & 64 & 0.0173 \\
    2 & 0.0155 & 128 & 0.0201 \\
    4 & 0.0152 & 256 & 0.0258 \\
    8 & 0.0153 & 512 & 0.0361 \\
    16 & 0.0151 & 1024 & 0.0600 \\
    32 & 0.0152 &  &  \\
    % Dendrite & Input terminal  & $\sim$100     \\
    % Axon     & Output terminal & $\sim$10      \\
    % Soma     & Cell body       & up to $10^6$  \\
    \bottomrule
  \end{tabular}
\end{table}
%%%%%%%%%%%%%%%%%% table 2 ends %%%%%%%%%%%%%%%%%%%%%%%%%%%%%%%%%%%%%%
Again, important components can be identified. The component with the maximum value of equation 3 is assumed to apply on the given input(X) as perturbation analysis may not work with large number of parameters. Equations 3 uses partial derivatives instead of perturbations to identify components.
\begin{equation} sum((\partial y(x,w)/\partial w)^2|x=X, \forall w  \in f(x,w))\end{equation}An example is shown in Figure 7. Most vectors in this plane are very close to each other thus a linear approximation can be made(Figure 7(a)). This is also verified by plotting the model output with respect to linear component output(Figure 7(c)) and by using linear regression to verify linearity(Figure 7(b)). Components can also be found using max-min form as log-exp form approximates it(refer Appendix B). This allows much easier identification and will not require any separate computation at all in cases described later. This method will be used in all following sections.
%%%%%%%%%%%%%%%%%%%% Figure 6(updated) %%%%%%%%%%%%%%%%%%%%%%%%%%%
\begin{figure}[ht]
  \centering
  \includegraphics[width=4in,height=1in]{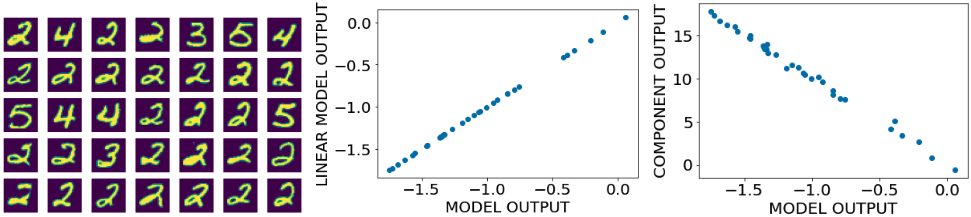}
%   \fbox{\rule[-.5cm]{0cm}{4cm} \rule[-.5cm]{4cm}{0cm}}
  \caption{Analysis of MNIST model component. Left(a): Points corresponding to a single component; Middle(b): Verification of linearity using linear regression; Right(c): Model output with respect to the expected component function output(Showing that set visualisations are retained even in higher dimensions). The slope is negative as the minus sign is absorbed in the learnt weight in the log-exp fomrulation.}
\end{figure}
%%%%%%%%%%%%%%%%%%% Figure 6 ends %%%%%%%%%%%%%%%%%%%%%%%%%
\subsection{As a classifier}
The classifier architecture is tested thoroughly on a simulated dataset and on benchmark datasets like MNIST, FMNIST and CIFAR 10 as shown:
\subsubsection{Simulated Data}

%%%%%%%%%%%%%%%%%% Figure 8 Simulated Classification Dataset %%%%%%%%%%%%
\begin{figure}[ht]
  \centering
  \includegraphics[width=4.5in,height=1in]{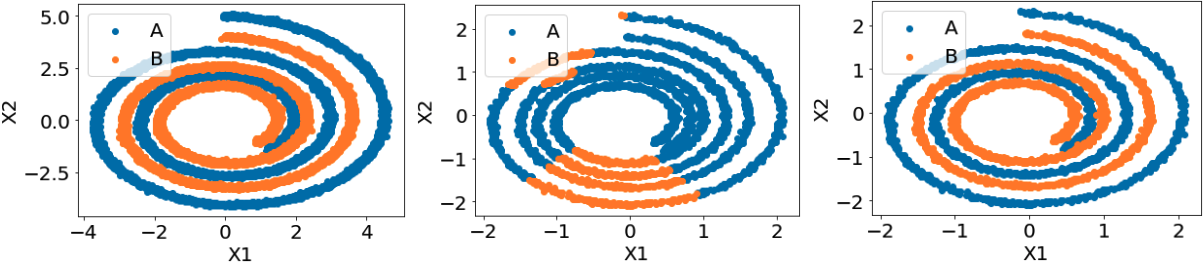}
%   \fbox{\rule[-.5cm]{0cm}{4cm} \rule[-.5cm]{4cm}{0cm}}
  \caption{Simulated Classification Dataset; Blue: Class A, Orange: Class B. Left: Actual Data; Middle: Before Training predictions on Test, Right: After training predictions on Test}\end{figure}
%%%%%%%%%%%%%%%% Figure ends %%%%%%%%%%%%%%%%%%%%%%%%%%%%%%
%%%%%%%%%%%%%%%%% Figure 9 %%%%%%%%%%%%%%%%%%%%%%%%%%%
\begin{figure}[ht]
  \centering
  \includegraphics[width=5in,height=1.5in]{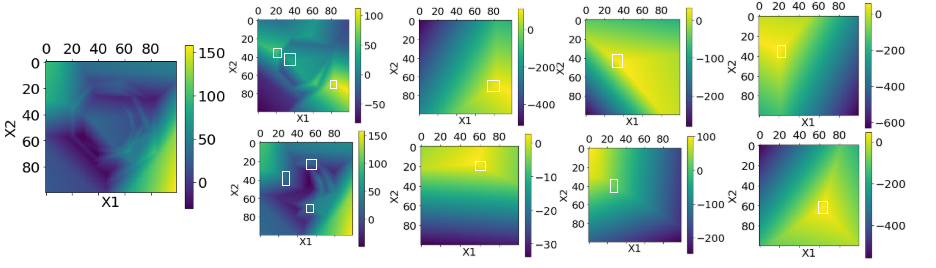}
%   \fbox{\rule[-.5cm]{0cm}{4cm} \rule[-.5cm]{4cm}{0cm}}
  \caption{Visualising Classifier. Colour represents z values. First Image: Decision Boundary using max instead of softmax; Row 1: regressor of classifier 1; Row 2: regressor of classifier 2. First image in each row represents the functions learnt by class regressors. Following images show major individual components being unioned to give the result. Boxes indicate easily identifiable features.}
\end{figure}
%%%%%%%%%%%%%% Figure 8 ends %%%%%%%%%%%%%%%%%%%%%%%%
A simulated spiral classification task with 2 classes and 10,000 examples per class is generated. These are then split into 80\% training and 20\% testing. The proposed classifier is tested and accuracy achieved is very high and comparable to FCNNs in [18]. Results are shown in Figure 8(more setup details in Appendix D). The simulated classification problem is also visualised in Figure 9. These visuals show the 3-dimensional function of the 2 regressors which are being used by each class. The individual intersected terms have edges and fork-shaped corners representing intersection of 3 planes. These are then unioned to generate the outputs of every regressor. The important features can be identified as shown. Then softmax is taken over classes which is represented using max instead to clarify the decision boundary.
\subsubsection{MNIST, FMNIST, CIFAR 10}
The architecture is also trained on benchmark datasets MNIST[19], FMNIST[20], and CIFAR 10[21]. 10 regressors with $u$ and $i$ values shown in Table 2 are used. 0.1 and 0.05 are used for $a$ in the modified variant of the model in CIFAR 10. The modified classifier is used for CIFAR 10 with the first linear layer reducing dimensions from 3072 to 300. Similar work was done in [22] on CIFAR 10 dataset with FCNNs. Log-exp form with sigmoidal components is also used and compared below. The result with sigmoid components has similar interpretations as Disjunctive Normal Networks and the individual classes are shown in Appendix M. The current experimentation did not use PCA whitening and has much lesser parameters thus the results were slightly worse than [22] on CIFAR 10 when their simple Relu version is considered. Yet, they are still comparable on other datasets as shown in Table 3. The neural networks shown are trained in the same environment (details and convergence graphs in Appendix I).

Simultaneously the log-exp layer was also used as the final layer with a CNN backbone which was not pretrained. This new architecture was trained on MNIST, FMNIST and CIFAR10 datasets and achieved a performance on par with fully connected layers. The mode achieved 99.52\% on MNIST, 92\% on FMNIST and 74\% on CIFAR10.

%%%%%%%%%%%%%%%% Table 2 %%%%%%%%%%%%%%%%%%%%%%%%%%%%
% \begin{table}[ht]
%   \caption{ Classification results on MNIST, FMNIST, CIFAR 10}
% %   \label{}
%   \centering
%   \begin{tabular}{llllllllll}
%     \toprule
%     \multicolumn{1}{c}{Dataset}                  
%     &
%     \multicolumn{3}{c}{Proposed Model}            
%     &
%     \multicolumn{3}{c}{1 Hidden Layer NN}            
%     &
%     \multicolumn{3}{c}{3 Hidden Layer NN}            \\
%     \cmidrule(r){1-10}
%      & Accuracy & Time & Epochs & Accuracy & Time & Epochs & Accuracy & Time & Epochs \\
%      & (u, i) & per & to &  & per & to &  & per & to \\
%      & values & epoch & converge &  & epoch & converge &  & epoch & converge \\
%      &  & (ms) & &  & (ms) & &  & (ms) & \\
%     \midrule
%     MNIST & 98 (16,4) & 1.5 & 10 & 98 & 105.3 & 8 & 98 & 128.4 & 12 \\
%     FMNIST & 89 (16,4) & 1.3 & 35 & 89 & 96.1 & 15 & 89 & 129.4 & 17 \\ 
%     CIFAR 10 & 51 (1,64) & 1.7 & 43 & 45 & 88.6 & 19 & 49 & 115.8 & 17 \\

%     \bottomrule
%   \end{tabular}
% \end{table}

\begin{table}[ht]
  \caption{ Classification results on MNIST, FMNIST, CIFAR 10 (A: Accuracy; AMERA:Average Minimum Epochs for obtaining Reported Accuracy)}
%   \label{}
  \centering
  \begin{tabular}{lllllllll}
    \toprule
    % \multicolumn{1}{c}{Dataset}                  
    &
    \multicolumn{2}{c}{Proposed Linear Model}
    &
    \multicolumn{2}{c}{Proposed Sigmoidal Model}
    &
    \multicolumn{2}{c}{2 Hidden Layer NN}            
    &
    \multicolumn{2}{c}{4 Hidden Layer NN}            \\
    \cmidrule(r){2-9}
    Dataset & A (u, i)  & AMERA & A & AMERA & A & AMERA & A & AMERA \\
    \midrule
    MNIST  & 98 (16,4) & 10.2 & 98 & 9 & 98 & 6.9 & 98 & 13.6 \\
    FMNIST & 89 (40,3) & 19.9 & 89 & 19 & 89 & 16.1 & 89 & 18.5 \\
    CIFAR 10 & 51 (1,64) & 35.5 & 52 & 38 & 44 & 22.4 & 49 & 23.3 \\

    \bottomrule
  \end{tabular}
\end{table}
%%%%%%%%%%%%%%%%%%%%%%% Table ends %%%%%%%%%%%%%%%%%%%%%%%%%%%%%%

\subsubsection{Testing on UCI Tabular Datasets}
The architecture is also tested thoroughly on UCI common datasets and the accuracy of the log-exp form and max-min form with log-exp pretraining are recorded as shown below. The details of this pretraining are described later in this paper.

%%%%%%%%%%%%%%%%%%%% table 1 Mimicking a Fully Connected Neural Network %%%%%%%%% 
\begin{table}[ht]
  \caption{Results on Common UCI Tabular Datasets}
%   \label{}
  \centering
  \begin{tabular}{llll}
    \cmidrule(r){1-4}
    Dataset & Log-exp Model & Max-min Model & Relu MLP\\
    \midrule
     Iris & 100\% & 100\% & 100\% \\
    Breast Cancer Wisconsin & 97\% & 96\% & 97\% \\
    Boston House Prices (Regression MSE) & 0.30 & 0.32 & 0.38 \\
Haberman’s Survival & 72\% & 72\% & 70\% \\
PIMA India & 75\% & 75\% & 71\% \\
Image Segmentation & 91\% & 91\% & 92\% \\
 Glass Classification & 83\% & 83\% & 75\% \\
Balance Scale & 98\% & 98\% & 99\% \\
Bank Note Authentication & 100\% & 100\% & 100\% \\
US Voting & 95\% & 95\% & 97\% \\
Mushroom Dataset & 100\% & 100\% & 100\% \\
Titanic & 78\% & 78\% & 78\% \\
Car Dataset & 98\% & 98\% & 100\% \\
    % Dendrite & Input terminal  & $\sim$100     \\
    % Axon     & Output terminal & $\sim$10      \\
    % Soma     & Cell body       & up to $10^6$  \\
    \bottomrule
  \end{tabular}
\end{table}
%%%%%%%%%%%%%%%%%% table 2 ends %%%%%%%%%%%%%%%%%%%%%%%%%%%%%%%%%%%%%%
%%%%%%%%%%%%%%%%%%%%%%% Table ends %%%%%%%%%%%%%%%%%%%%%%%%%%%%%%

\section{Analysis of set structures}

A variety of set structures are implemented during experimentation. The main observations are qualitative in nature. Functions using purely unions or intersections did not perform well. Functions with a nearly equal number of unions and intersections worked the best and functions with slightly more unions than intersections generally performing better than its inverse. This behaviour is data and problem-dependent and is shown in Figure 3, Table 1 and Appendix D.

\section{Other Major Benefits:}

\subsection{Going back to the max-min form}

%%%%%%%%%%%%%%% Figure 9 Process of optimizing max-min form %%%%%%%%%%%%
\begin{figure}[ht]
  \centering
  \includegraphics[width=2.5in,height=1in]{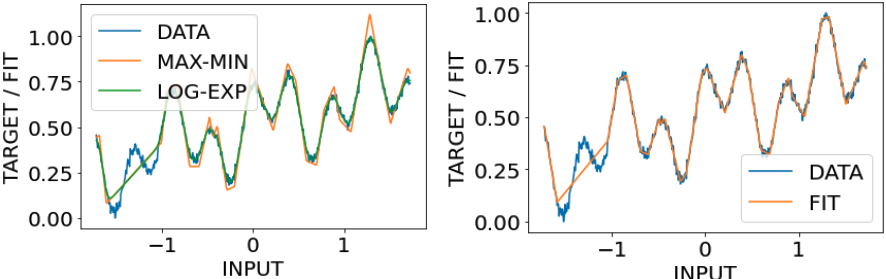}
%   \fbox{\rule[-.5cm]{0cm}{4cm} \rule[-.5cm]{4cm}{0cm}}
  \caption{Process of optimizing max-min form Left: Before fine-tuning, Blue: Data, Green:log-exp form, Yellow: max-min form Right: max-min form after fine-tuning}
\end{figure}
%%%%%%%%%%%%% Figure 9 ends %%%%%%%%%%%%%%%%%%%%%%%%

The max-min form is much simpler than the log-exp form but cannot be trained directly. This may be because new non-linearities are not introduced as the gradients with respect to unused parameters are zero as shown in Appendix E. Thus the log-exp form is trained first with linear components and the corresponding weights and biases are then used in the max-min form. The max-min form is then fine-tuned on the dataset to reduce the error. This results in a faster and simpler but slightly less accurate model as shown in Figure 10. Parameter $a$ should also take values in the range 5 to 20 as this allows better approximation of max-min form. This was tested on MNIST, FMNIST and CIFAR 10 as shown in Table 3. Similar behaviour is expected for non-linear components too but is not tested. 

%%%%%%%%%%%%%%%%%%% Table 4 %%%%%%%%%%%%%%%%%%%%%%%%%%
\begin{table}[ht]
  \caption{Fine-tuning max-min form on datasets. 20 Fine-tuning epochs were used. CIFAR 10 model was trained for 30 epochs.}
%   \label{}
  \centering
  \begin{tabular}{lll}
    % \cmidrule(r){1-3}
    \toprule
    Dataset (u, i)    & Accuracy of max-min form    & Accuracy of log-exp form\\
    \midrule
    MNIST (16, 4) & 97\% & 98\% \\
    FMNIST (16, 4) & 88\% & 88\% \\
    CIFAR 10 (25, 3, a=10) & 46\% & 45\% \\
    \bottomrule
  \end{tabular}
\end{table}
%%%%%%%%%%%%%%%%%% Table 4 ends %%%%%%%%%%%%%%%%%%%%%%
\subsection{Targeted Learning}

%A very strong implication of the independence of the weights is that they can be treated separately for individual components. This allows insertion of non-linearities in specific regions of the domain. To do this, a model with initially fewer(or with an unexpectedly small number) weights is trained on the dataset. This model has a higher loss that could be reduced by adding more weights. An analysis of the model will reveal which regions of the domain correspond to higher error. New weights can be created by finding the weights corresponding to those regions and adding random perturbations to them. All weights of a single intersection unit should be selected at once. These can then be perturbed and appended to the weight tensor, the same is done with biases and the model can be fine-tuned on the data. This specific injection of the weights and biases will allow the model to increase the amount of non-linearity for the region selected and thus reduce the error if possible. This was performed on the simulated regression dataset and demonstrated in Fig. 10.

Models, when trained on highly non-linear datasets, might not be able to fit well to all non-linearities in every part of the domain. To handle this, new components can be introduced in regions where the fit is inaccurate thus allowing more parameters to be used to fit accurately. For introducing new components, regions of inaccurate fit are identified by model analysis, the components initially being used there are then identified as shown previously. All intersection units using these components are identified. All component functions in these intersection units are replicated and small uniform random noise is added to every replication. These are then added to the model parameters. The model is then fine-tuned again on the dataset and the newly introduced components should now be able to approximate the function better. This is demonstrated in Figure 11. This is an important implication of the use of set operations.

%%%%%%%%%%%%%%%%%%%% Figure 10 Performing targeted learning %%%%%%%%%%%%%
\begin{figure}[ht]
  \centering
  \includegraphics[width=3in,height=1in]{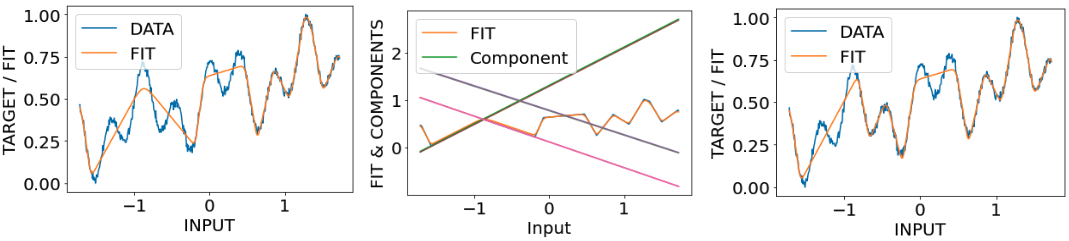}
%   \fbox{\rule[-.5cm]{0cm}{4cm} \rule[-.5cm]{4cm}{0cm}}
  \caption{Performing targeted learning
Left: Previously trained architecture Middle: Weights and perturbed weights Right: Improved fit}
\end{figure}
%%%%%%%%%%%%%% Figure 10 ends %%%%%%%%%%%%%%%%%%%%%%%

\section{Conclusion and Future Work}

It is shown that piecewise functions can be approximated and fit with similar performance using simple set operations and nested multi-layered architectures are not always needed. New directions like sharing of polytopes between classes(Appendix H) and piecewise combinations(Appendix G) are interesting for exploration, and directions like “union” and “intersection” of pretrained models can be interesting pursuits from here. Targeted learning may be extended to more realistic datasets and handle major challenges like model bias and concept drifts. The current analysis is not application-specific thus more detailed and thorough field-specific analysis of the architecture with CNNs, RNNs and many other designs will be useful as well. The log-exp form is differentiable and simple thus can find applications in other scientific and numeric fields. Many more research avenues may open to study transfer, unsupervised and reinforcement learning applications using the presented architectures. 

\section*{Broader Impact}

\subsection*{Major Impact:}

This paper fundamentally aims at providing an architecture which is much more intuitive and easy to design, debug and analyse. These properties are sought for in many applications ranging from Neural Network based controllers to complex computer vision and natural language tasks to theoretical fields related to AI and optimisation to system modelling. Most sections in the paper have demonstrated new avenues and directions which are a direct outcome of use of set operations instead of multi-layered or nested architectures.

 High interpretability and ease of use should allow SPINEs to be tested in visual and language systems instead of fully connected layers. More approaches for training akin to targeted learning and ones based on polytope manipulations can also find applications in this formulation. The model can also be initialised to special better understood states, which enables ease of combining human knowledge and machine learning. Essential extensions to these ideas for better adoption are also provided in the Appendix. These include mixing of different piecewise functions(for situations with prior knowledge regarding the data) and shared polytopes in classifiers(to control the growth of parameters with classes and for applications in fields like hierarchical classification).

We strongly believe in "build what you know or know what you built". In deep learning, a huge focus was on the latter half and amazing interpretations for neurons, layers, features and possible failures were seen in a plethora of interpretability papers. This paper might have scratched the surface of the first half and hopes it has presented a case for digging deeper into it. Better understood architectures amount to better predictability of behaviour, better understanding and reasoning for functions learnt and more importantly increased trust in the system. It is also believed that better understood architectures may be able to follow important standards and regulations better and closer thus this can be an important direction

\subsection*{Possible improvements to this work:}

It is also realised that the theoretical foundations for the presented design can go deeper further. The presented idea will automatically cause more transfusion of ideas between fields like tropical geometry, optimisation and different aspects of machine and deep learning. The observation that the presented architecture performs equally well with much lesser components than FCNNs is thought-provoking as well. Challenges were also faced with exponentiation and large numbers during implementation and more solutions to handle them are needed. A tradeoff between interpretability(larger $a$) and ease of training(smaller $a$ allows larger gradients for unexpressed components) was also seen and may open avenues to formulations with learnable $a$ or other ways to handle it. Capabilities of non-linear components can also be explored on more realistic datasets as current resources did not permit more extensive analysis. Exploration of the given architecture in many different contexts like when used with CNNs and RNNs is also needed. This work provides a proof of concept for many ideas like Targeted Learning and Polytope sharing, but these have to be tested on larger datasets and in more practical settings.

\subsection*{Ethical Concerns:}

Our major concerns are with Targeted Learning and other means to modify a well-trained model. The simplicity of design also allows the ability to intentionally introduce biases or faults in the model which might not be detected by using simple tests. It will be important to track the number and some metadata regarding the polytopes and how they are being used. An adversary might introduce new polytopes or shift the existing ones because the effects of such attacks are much more predictable now. The simplicity of design may also be a solution to such attacks. The tampered model can be compared with the original model if the original version is available or, properties like the number of polytopes and other metadata can be checked and tracked to detect tampering. For correcting such malicious modifications, the model can undergo fine-tuning or targeted learning or the modified or added polytopes can be detected using the above checks and handled accordingly. 

\subsection*{Concluding Remarks:}
More research in this direction is surely needed and it shows amazing prospects of fruitfulness and providing benefits in many real life cases. We hope SPINE is added to the spinal-chord of deep learning, giving it more interpretability and intuitiveness.

\section*{Acknowledgements:}

This research is not funded by any organisation and is done using only open source or freely available tools. We thank Google Colaboratory, Desmos and Geogebra for providing free platforms for creating models and visualisations. Special credits go to all anonymous reviewers for their time and essential inputs.

\section*{References}

% References follow the acknowledgments. Use unnumbered first-level heading for
% the references. Any choice of citation style is acceptable as long as you are
% consistent. It is permissible to reduce the font size to \verb+small+ (9 point)
% when listing the references.
% {\bf Note that the Reference section does not count towards the eight pages of content that are allowed.}
\medskip

\begin{small}

[1] He, K., Zhang, X., Ren, S., \& Sun, J.\ (2016). Deep residual learning for image recognition. In Proceedings of the IEEE conference on computer vision and pattern recognition (pp. 770-778).

[2] Howard, J., \& Ruder, S.\ (2018). Universal language model fine-tuning for text classification. arXiv preprint arXiv:1801.06146.

[3] Kao, H. C., Tang, K. F., \& Chang, E. Y.\ (2018, April). Context-aware symptom checking for disease diagnosis using hierarchical reinforcement learning. In Thirty-Second AAAI Conference on Artificial Intelligence.

[4] Smilkov, D., Thorat, N., Kim, B., Viégas, F., \& Wattenberg, M.\ (2017). Smoothgrad: removing noise by adding noise. arXiv preprint arXiv:1706.03825.

[5] Selvaraju, R. R., Cogswell, M., Das, A., Vedantam, R., Parikh, D., \& Batra, D.\ (2017). Grad-cam: Visual explanations from deep networks via gradient-based localization. In Proceedings of the IEEE international conference on computer vision (pp. 618-626).

[6] Ribeiro, M. T., Singh, S., \& Guestrin, C.\ (2016, August). “ Why should i trust you?” Explaining the predictions of any classifier. In Proceedings of the 22nd ACM SIGKDD international conference on knowledge discovery and data mining (pp. 1135-1144).

[7] Mahendran, A., \& Vedaldi, A.\ (2015). Understanding deep image representations by inverting them. In Proceedings of the IEEE conference on computer vision and pattern recognition (pp. 5188-5196).

[8] Li, H., Xu, Z., Taylor, G., Studer, C., \& Goldstein, T.\ (2018). Visualizing the loss landscape of neural nets. In Advances in Neural Information Processing Systems (pp. 6389-6399).

[9] Serra, T., Tjandraatmadja, C., \& Ramalingam, S.\ (2017). Bounding and counting linear regions of deep neural networks. arXiv preprint arXiv:1711.02114.

[10] Hanin, B., \& Sellke, M.\ (2017). Approximating continuous functions by relu nets of minimal width. arXiv preprint arXiv:1710.11278.

[11] Leshno, M., Lin, V. Y., Pinkus, A., \& Schocken, S.\ (1993). Multilayer feedforward networks with a nonpolynomial activation function can approximate any function. Neural networks, 6(6), 861-867.

[12] Grover, J. S.\ (2019). Differentiable Set Operations for Algebraic Expressions. arXiv preprint arXiv:1912.12181.

[13] https://www.desmos.com/calculator/12zjzrtfwb

[14] https://www.geogebra.org/3d/jabt3tw7,Sandall, B. (2015). iPad/Tablet/Computer application: GeoGebra3D. Mathematics and Computer Education, 49(2), 149.

[15] Ovchinnikov, S.\ (2002). Max-min representation of piecewise linear functions. Contributions to Algebra and Geometry, 43(1), 297-302.

[16] G. C. Calafiore, S. Gaubert and C. Possieri, "Log-Sum-Exp Neural Networks and Posynomial Models for Convex and Log-Log-Convex Data," in IEEE Transactions on Neural Networks and Learning Systems, vol. 31, no. 3, pp. 827-838, March 2020, doi: 10.1109/TNNLS.2019.2910417.

[17] G. C. Calafiore, S. Gaubert and C. Possieri, "A Universal Approximation Result for Difference of Log-Sum-Exp Neural Networks," in IEEE Transactions on Neural Networks and Learning Systems, doi: 10.1109/TNNLS.2020.2975051.

[18] Tensorflow, P.\ (2017). A Neural Network Playground.

[19] LeCun, Y., Bottou, L., Bengio, Y., \& Haffner, P.\ (1998). Gradient-based learning applied to document recognition. Proceedings of the IEEE, 86(11), 2278-2324.

[20] Xiao, H., Rasul, K., \& Vollgraf, R.\ (2017). Fashion-mnist: a novel image dataset for benchmarking machine learning algorithms. arXiv preprint arXiv:1708.07747.

[21] Krizhevsky, A., Nair, V., \& Hinton, G.\ (2009). CIFAR-10 (canadian institute for advanced research).(2009). URL http://www. cs. toronto. edu/kriz/cifar. html.

[22] Lin, Z., Memisevic, R., \& Konda, K.\ (2015). How far can we go without convolution: Improving fully-connected networks. arXiv preprint arXiv:1511.02580.

[23] Wan, Alvin, \& Lisa Dunlap, \& Daniel Ho, \& Jihan Yin, \& Scott Lee, \& Henry Jin, \& Suzanne Petryk, \& Sarah Adel Bargal, and Joseph E. Gonzalez. "NBDT: Neural-backed decision trees." arXiv preprint arXiv:2004.00221 (2020).

[24] Yang, Yongxin, \& Irene Garcia Morillo, and Timothy M. Hospedales. "Deep neural decision trees." arXiv preprint arXiv:1806.06988 (2018).

[25] Seyedhosseini, \& Mojtaba, \& Mehdi Sajjadi, and Tolga Tasdizen. "Image segmentation with cascaded hierarchical models and logistic disjunctive normal networks." Proceedings of the IEEE international conference on computer vision. 2013.

[26] Veness, Joel, \& Tor Lattimore, \& Avishkar Bhoopchand, \& David Budden, \& Christopher Mattern, \& Agnieszka Grabska-Barwinska, \& Peter Toth, \& Simon Schmitt, and Marcus Hutter. "Gated linear networks." arXiv preprint arXiv:1910.01526 (2019).

\end{small}

\newpage

% \newpage
\appendixpage
\tableofcontents
\newpage
\appendix
\section{Derivation of the Log-exp form using set operations}

\subsection{Important Concepts and Definitions}
As stated in the paper, piecewise linear functions can be expressed as a union of several polytopes or other concave piecewise linear functions and these can in turn be expressed as an intersection of various linear inequations. This implies that every piecewise linear function is expressed as a combination of set operations(unions and intersections) applied on sets which are represented as linear inequations. Thus a set here can be defined as follows:

$A\textsubscript{ij}=\{ (x,y)| x$ and $y$ satisfy the inequality $y <=w\textsuperscript{T}\textsubscript{i,j}x+b\textsubscript{i,j}$;where $x,w \in R\textsuperscript{n}$ and $y,b \in R$, and $i,j$ are indices \}

Conjunctive and Disjunctive Normal Forms are general representations for all combinations of set operations thus a general representation of piecewise linear functions can be obtained using either of these. Conversion of set operations to these forms may also involve negations which we assume are learnt by component functions(by negating some parameters) and thus we do not explicitly handle them.

As stated in the paper, empirically, the max-min forms do not train well and this is shown in Appendix E. Thus, for deriving the log-exp form, the set operations can be approximated using the transforms in [1] (Refer Appendix References). These include the following:

\begin{enumerate}
  \item \textbf{Set representation:} The sets are initially represented as $f(x,y)<=0$ and then are first converted to the form $e\textsuperscript{f(x,y)}<=1$
   \item \textbf{Negation:} Negation of set operations is represented as $e\textsuperscript{-f(x,y)}<=1$
   \item \textbf{Intersection:} Intersection of these sets is approximated as e\textsuperscript{af\textsubscript{1}(x,y)}+e\textsuperscript{af\textsubscript{2}(x,y)}+e\textsuperscript{af\textsubscript{3}(x,y)}+...<=1
   \item \textbf{Union:} Union of these sets is approximated as (e\textsuperscript{-af\textsubscript{1}(x,y)}+e\textsuperscript{-af\textsubscript{2}(x,y)}+e\textsuperscript{-af\textsubscript{3}(x,y)}+...)\textsuperscript{-1}<=1
   \item \textbf{Nested set operations:} When multiple such operations are nested, then e\textsuperscript{af(x,y)}can be replaced by the LHS of the resultant of the previous set operations
\end{enumerate}

These results are derived and detailed in [1].

\subsection{Proving using the disjunctive normal form(DNF)}

Let there be m sets being intersected in every term and n such terms, then the disjunctive normal form can be written as:

$S=(A\textsubscript{1,1}\cap A\textsubscript{1,2} \cap A\textsubscript{1,3} \cap ... A\textsubscript{1,m}) \cup (A\textsubscript{2,1}\cap A\textsubscript{2,2} \cap A\textsubscript{2,3} \cap ... A\textsubscript{2,m}) \cup (A\textsubscript{3,1}\cap A\textsubscript{3,2} \cap A\textsubscript{3,3} \cap ... A\textsubscript{3,m})... (A\textsubscript{n,1}\cap A\textsubscript{n,2} \cap A\textsubscript{n,3} \cap ... A\textsubscript{n,m})$

When the intersection is performed on inequalities of the form $y\textsubscript{ij}<=w\textsuperscript{T}\textsubscript{ij}x+b\textsubscript{ij}$, then the resultant inequality will take the form $y\textsubscript{i}<=min(w\textsuperscript{T}\textsubscript{ij}x+b\textsubscript{ij}) \exists i \forall j$ which is pointwise min, and when union is performed on the results which is pointwise max of these intersection operations, then the result will take the form $y<=max(min(w\textsuperscript{T}\textsubscript{ij}x+b\textsubscript{ij}) \exists i \forall j) \forall i$. This is the max-min form in the paper. It is more rigorously derived and proven in [2] as well.

For deriving the log-exp form , the sets are represented as $y\textsubscript{ij}-(w\textsubscript{ij}\textsuperscript{T}x+b\textsubscript{ij})<=0$ (for linear components) thus obtaining the form $e\textsuperscript{a(y\textsubscript{ij}-(w\textsubscript{ij}\textsuperscript{T}x+b\textsubscript{ij}))}<=1$. Intersection operations can then be applied as:

$e\textsuperscript{a(y\textsubscript{i,1}-(w\textsubscript{i,1}\textsuperscript{T}x+b\textsubscript{i,1}))} + e\textsuperscript{a(y\textsubscript{i,2}-(w\textsubscript{i,2}\textsuperscript{T}x+b\textsubscript{i,2}))} + e\textsuperscript{a(y\textsubscript{i,3}-(w\textsubscript{i,3}\textsuperscript{T}x+b\textsubscript{i,3}))} + ... e\textsuperscript{a(y\textsubscript{i,m}-(w\textsubscript{i,m}\textsuperscript{T}x+b\textsubscript{i,m}))} <= 1$

This is followed by union operations as:

((e\textsuperscript{a(y\textsubscript{1,1}-(w\textsubscript{1,1}\textsuperscript{T}x+b\textsubscript{1,1}))} + e\textsuperscript{a(y\textsubscript{1,2}-(w\textsubscript{1,2}\textsuperscript{T}x+b\textsubscript{1,2}))} + e\textsuperscript{a(y\textsubscript{1,3}-(w\textsubscript{1,3}\textsuperscript{T}x+b\textsubscript{1,3}))} + ... e\textsuperscript{a(y\textsubscript{1,m}-(w\textsubscript{1,m}\textsuperscript{T}x+b\textsubscript{1,m}))})\textsuperscript{-1} + 

(e\textsuperscript{a(y\textsubscript{2,1}-(w\textsubscript{2,1}\textsuperscript{T}x+b\textsubscript{2,1}))} + e\textsuperscript{a(y\textsubscript{2,2}-(w\textsubscript{2,2}\textsuperscript{T}x+b\textsubscript{2,2}))} + e\textsuperscript{a(y\textsubscript{2,3}-(w\textsubscript{2,3}\textsuperscript{T}x+b\textsubscript{2,3}))} + ... e\textsuperscript{a(y\textsubscript{2,m}-(w\textsubscript{2,m}\textsuperscript{T}x+b\textsubscript{2,m}))})\textsuperscript{-1} + 

(e\textsuperscript{a(y\textsubscript{3,1}-(w\textsubscript{3,1}\textsuperscript{T}x+b\textsubscript{3,1}))} + e\textsuperscript{a(y\textsubscript{3,2}-(w\textsubscript{3,2}\textsuperscript{T}x+b\textsubscript{3,2}))} + e\textsuperscript{a(y\textsubscript{3,3}-(w\textsubscript{3,3}\textsuperscript{T}x+b\textsubscript{3,3}))} + ... e\textsuperscript{a(y\textsubscript{3,m}-(w\textsubscript{3,m}\textsuperscript{T}x+b\textsubscript{3,m}))})\textsuperscript{-1} + ...

(e\textsuperscript{a(y\textsubscript{n,1}-(w\textsubscript{n,1}\textsuperscript{T}x+b\textsubscript{n,1}))} + e\textsuperscript{a(y\textsubscript{n,2}-(w\textsubscript{n,2}\textsuperscript{T}x+b\textsubscript{n,2}))} + e\textsuperscript{a(y\textsubscript{n,3}-(w\textsubscript{n,3}\textsuperscript{T}x+b\textsubscript{n,3}))} + ... e\textsuperscript{a(y\textsubscript{n,m}-(w\textsubscript{n,m}\textsuperscript{T}x+b\textsubscript{n,m}))})\textsuperscript{-1})\textsuperscript{-1} <= 1

$ \implies (\sum_{i=1}^{n}(\sum_{j=1}^{m}e\textsuperscript{a(y\textsubscript{i,j}-(w\textsuperscript{T}\textsubscript{i,j}x+b\textsubscript{i,j}))})\textsuperscript{-1})\textsuperscript{-1} <= 1 $

Taking only the terms which lie on the boundary gives:

$ \implies (\sum_{i=1}^{n}(\sum_{j=1}^{m}e\textsuperscript{a(y\textsubscript{i,j}-(w\textsuperscript{T}\textsubscript{i,j}x+b\textsubscript{i,j}))})\textsuperscript{-1})\textsuperscript{-1} = 1 $

$ \implies (\sum_{i=1}^{n}(\sum_{j=1}^{m}e\textsuperscript{a(y\textsubscript{i,j}-(w\textsuperscript{T}\textsubscript{i,j}x+b\textsubscript{i,j}))})\textsuperscript{-1})\textsuperscript{-1} = 1 $

$ \implies e\textsuperscript{ay}(\sum_{i=1}^{n}(\sum_{j=1}^{m}e\textsuperscript{-a(w\textsuperscript{T}\textsubscript{i,j}x+b\textsubscript{i,j})})\textsuperscript{-1})\textsuperscript{-1} = 1 $

$ \implies \sum_{i=1}^{n}(\sum_{j=1}^{m}e\textsuperscript{-a(w\textsuperscript{T}\textsubscript{i,j}x+b\textsubscript{i,j})})\textsuperscript{-1} = e\textsuperscript{ay} $

$ \implies y=\ln(\sum_{i=1}^{n}(\sum_{j=1}^{m}e\textsuperscript{-a(w\textsuperscript{T}\textsubscript{i,j}x+b\textsubscript{i,j})})\textsuperscript{-1})/a$

$ \implies y=\ln(\sum_{i=1}^{n}1/\sum_{j=1}^{m}e\textsuperscript{-a(w\textsuperscript{T}\textsubscript{i,j}x+b\textsubscript{i,j})})/a$

Extending this result to non-linear continuous components which are represented by 
y\textsubscript{i,j}-f\textsubscript{i,j}(x,$\theta$) <= 0 gives:

%$ \implies y=\ln(\sum_{i=1}^{n}1/\sum_{j=1}^{m}e\textsuperscript{-a(f\textsubscript{i,j}(x, \theta)})/a$

 \(  \implies y=ln \left(  \sum _{i=1}^{n}1/ \sum _{j=1}^{m}e^{-af_{\text{i, j}} \left( x, \theta  \right) } \right) /a \)

\subsection{Proving using the conjunctive normal form(CNF)}

Let there be m sets being unioned in every term and n such terms, then the conjunctive normal form can be written as:

$S=(A\textsubscript{1,1}\cup A\textsubscript{1,2} \cup A\textsubscript{1,3} \cup ... A\textsubscript{1,m}) \cap (A\textsubscript{2,1}\cup A\textsubscript{2,2} \cup A\textsubscript{2,3} \cup ... A\textsubscript{2,m}) \cap (A\textsubscript{3,1}\cup A\textsubscript{3,2} \cup A\textsubscript{3,3} \cup ... A\textsubscript{3,m})... (A\textsubscript{n,1}\cup A\textsubscript{n,2} \cup A\textsubscript{n,3} \cup ... A\textsubscript{n,m})$

When the union is performed on inequalities of the form $y\textsubscript{ij}<=w\textsuperscript{T}\textsubscript{ij}x+b\textsubscript{ij}$, then the resultant inequality will take the form $y\textsubscript{i}<=max(w\textsuperscript{T}\textsubscript{ij}x+b\textsubscript{ij}) \exists i \forall j$ which is pointwise max, and when the intersection is performed on the results of these union operations, then the result will take the form $y<=min(max(w\textsuperscript{T}\textsubscript{ij}x+b\textsubscript{ij}) \exists i \forall j) \forall i$ which is pointwise min of the outputs of pointwise max. This form is equivalent to max-min form of the paper and can be used alternatively but has a slightly different intuition.

For deriving the log-exp form, the sets are represented as: $y\textsubscript{ij}-(w\textsubscript{ij}\textsuperscript{T}x+b\textsubscript{ij})<=0$ (for linear components) thus obtaining the form $e\textsuperscript{a(y\textsubscript{ij}-(w\textsubscript{ij}\textsuperscript{T}x+b\textsubscript{ij}))}<=1$. Union operations can then be applied as:

$(e\textsuperscript{-a(y\textsubscript{i,1}-(w\textsubscript{i,1}\textsuperscript{T}x+b\textsubscript{i,1}))} + e\textsuperscript{-a(y\textsubscript{i,2}-(w\textsubscript{i,2}\textsuperscript{T}x+b\textsubscript{i,2}))} + e\textsuperscript{-a(y\textsubscript{i,3}-(w\textsubscript{i,3}\textsuperscript{T}x+b\textsubscript{i,3}))} + ... e\textsuperscript{-a(y\textsubscript{i,m}-(w\textsubscript{i,m}\textsuperscript{T}x+b\textsubscript{i,m}))})\textsuperscript{-1} <= 1$

This is followed by intersection operations as:

(e\textsuperscript{-a(y\textsubscript{1,1}-(w\textsubscript{1,1}\textsuperscript{T}x+b\textsubscript{1,1}))} + e\textsuperscript{-a(y\textsubscript{1,2}-(w\textsubscript{1,2}\textsuperscript{T}x+b\textsubscript{1,2}))} + e\textsuperscript{-a(y\textsubscript{1,3}-(w\textsubscript{1,3}\textsuperscript{T}x+b\textsubscript{1,3}))} + ... e\textsuperscript{-a(y\textsubscript{1,m}-(w\textsubscript{1,m}\textsuperscript{T}x+b\textsubscript{1,m}))})\textsuperscript{-1} + 

(e\textsuperscript{-a(y\textsubscript{2,1}-(w\textsubscript{2,1}\textsuperscript{T}x+b\textsubscript{2,1}))} + e\textsuperscript{-a(y\textsubscript{2,2}-(w\textsubscript{2,2}\textsuperscript{T}x+b\textsubscript{2,2}))} + e\textsuperscript{-a(y\textsubscript{2,3}-(w\textsubscript{2,3}\textsuperscript{T}x+b\textsubscript{2,3}))} + ... e\textsuperscript{-a(y\textsubscript{2,m}-(w\textsubscript{2,m}\textsuperscript{T}x+b\textsubscript{2,m}))})\textsuperscript{-1} + 

(e\textsuperscript{-a(y\textsubscript{3,1}-(w\textsubscript{3,1}\textsuperscript{T}x+b\textsubscript{3,1}))} + e\textsuperscript{-a(y\textsubscript{3,2}-(w\textsubscript{3,2}\textsuperscript{T}x+b\textsubscript{3,2}))} + e\textsuperscript{-a(y\textsubscript{3,3}-(w\textsubscript{3,3}\textsuperscript{T}x+b\textsubscript{3,3}))} + ... e\textsuperscript{-a(y\textsubscript{3,m}-(w\textsubscript{3,m}\textsuperscript{T}x+b\textsubscript{3,m}))})\textsuperscript{-1} + ...

(e\textsuperscript{-a(y\textsubscript{n,1}-(w\textsubscript{n,1}\textsuperscript{T}x+b\textsubscript{n,1}))} + e\textsuperscript{-a(y\textsubscript{n,2}-(w\textsubscript{n,2}\textsuperscript{T}x+b\textsubscript{n,2}))} + e\textsuperscript{-a(y\textsubscript{n,3}-(w\textsubscript{n,3}\textsuperscript{T}x+b\textsubscript{n,3}))} + ... e\textsuperscript{-a(y\textsubscript{n,m}-(w\textsubscript{n,m}\textsuperscript{T}x+b\textsubscript{n,m}))})\textsuperscript{-1} <= 1

$ \implies \sum_{i=1}^{n}(\sum_{j=1}^{m}e\textsuperscript{-a(y\textsubscript{i,j}-(w\textsuperscript{T}\textsubscript{i,j}x+b\textsubscript{i,j})}))\textsuperscript{-1} <= 1 $

Taking only the terms which lie on the boundary gives:

$ \implies \sum_{i=1}^{n}(\sum_{j=1}^{m}e\textsuperscript{-a(y-(w\textsuperscript{T}\textsubscript{i,j}x+b\textsubscript{i,j}))})\textsuperscript{-1} = 1 $

$ \implies e\textsuperscript{ay}\sum_{i=1}^{n}(\sum_{j=1}^{m}e\textsuperscript{a(w\textsuperscript{T}\textsubscript{i,j}x+b\textsubscript{i,j})})\textsuperscript{-1} = 1 $

$\implies \sum_{i=1}^{n}(\sum_{j=1}^{m}e\textsuperscript{a(w\textsuperscript{T}\textsubscript{i,j}x+b\textsubscript{i,j})})\textsuperscript{-1} = e\textsuperscript{-ay}$

$\implies y = -\ln({\sum_{i=1}^{n}(\sum_{j=1}^{m}e\textsuperscript{a(w\textsuperscript{T}\textsubscript{i,j}x+b\textsubscript{i,j})})\textsuperscript{-1}})/a$

$\implies y = -\ln({\sum_{i=1}^{n}(1/\sum_{j=1}^{m}e\textsuperscript{a(w\textsuperscript{T}\textsubscript{i,j}x+b\textsubscript{i,j})})})/a$

Extending this result to non-linear continuous components which are represented by 
y\textsubscript{i,j}-f\textsubscript{i,j}(x,$\theta$) <= 0 gives:

\(  \implies y=-ln \left(  \sum _{i=1}^{n}1/ \sum _{j=1}^{m}e^{af_{\text{i, j}} \left( x, \theta  \right) } \right) /a \)

\subsection{Intuitions as a Neural Architecture}

The log-exp form as stated is an approximated piecewise function with very strong similarities to neural networks. Softmax and Softplus also take this form and even some loss functions do so. The architecture is trained in PyTorch using Adam optimizer with MSE and negative log loss just like neural networks. Similar performances were also achieved as shown in the paper and other sections of this Appendix. Yet, we do not consider log-exp form as just another activation function because it is able to approximate very complex functions as shown. This also means that nesting and making multiple-layers of this function is not needed(at least for fully connected structures) thus retaining a very large part of its interpretability and intuitiveness. It is much more amenable to consider the components and inequalities as "neurons" of this architecture and the two forms as "organisation"(neurons are expressed where needed) of neurons than as a "network"(neurons don't feed output into other neurons) of neurons. This makes explaining and analysing other sections and important ideas like sharing of polytopes (Appendix H) and mixture of component functions (Appendix G) much easier.

\section{The log-exp form tends to max(min(f(x, \texorpdfstring{$\theta$}))) as ‘a’ approaches infinity:}
The log-exp form is an approximation of max(min(f(x, $\theta$))) which can be proven by taking the limit on the parameter ‘a’ to tend to infinite.

To Prove:

 \( \mathop{\lim }_{a \rightarrow \infty} \left(  ln \left(  \sum _{i=1}^{n}1/ \left(  \sum _{j=1}^{m}e^{-afi,j \left( x, \theta  \right) } \right)  \right) /a  \right) =max \left( min \left( f(x, \theta) \right)   \exists i~ \forall j \right)   \forall i \)

Thus:

 \( \mathop{\lim }_{a \rightarrow \infty} \left(  ln \left(  \sum _{i=1}^{n}1/ \left(  \sum _{j=1}^{m}e^{-afi,j \left( x, \theta  \right) } \right)  \right) /a  \right)  \)
 
 \( \mathop{\lim }_{a \rightarrow \infty} \left(  ln \left(   \left(  \sum _{i=1}^{n}1/ \left(  \sum _{j=1}^{m}e^{-afi,j \left( x, \theta  \right) } \right)  \right) ^{1/a} \right)   \right)  \)
 
  \( \mathop{\lim }_{a \rightarrow \infty} \left(  ln \left(   \left(  \sum _{i=1}^{n}e^{afi,min \left( x \right) }/ \left(  \sum _{j=1}^{m}e^{-a \left( fi,j \left( x \right) -fi,min \left( x \right)  \right) } \right)  \right) ^{1/a} \right) ~ \right)\)
  
  ; taking min with respect to j common from inner summation 
  
  \( \mathop{\lim }_{a \rightarrow \infty} \left(  ln \left(   \left( e^{afmax,min \left( x \right) }\ast \sum _{i=1}^{n}e^{a \left( fi,min \left( x \right) -fmax,min \left( x \right)  \right) }/ 
  \left(  \sum _{j=1}^{m}e^{-a \left( fi,j \left( x \right) -fi,min \left( x \right)  \right) } \right)  \right) ^{1/a} \right) ~ \right) \); taking max of mins with respect to i common 
  
  \( f_{max,min} \left( x \right) +\mathop{\lim }_{a \rightarrow \infty} \left(  ln \left(   \left(  \sum _{i=1}^{n}e^{a \left( fi,min \left( x \right) -fmax,min \left( x \right)  \right) }/ \left(  \sum _{j=1}^{m}e^{-a \left( fi,j \left( x \right) -fi,min \left( x \right)  \right) } \right)  \right) ^{1/a} \right) ~ \right)  \) ~~~  ; As a is cancelled fmax,min \(\left( x \right)\)  can be taken out 

\( f_{max,min} \left( x \right) +\mathop{\lim }_{a \rightarrow \infty} (  ln \left(  \sum _{i=1}^{n}1/ \left(  \sum _{j=1}^{m}e^{a \left( fmax,min \left( x \right) -fi,j \left( x \right)  \right) } \right) /a~ \right) \) ; Rearranging 
  
  f\textsubscript{max,min}(x)+0 ; the limit takes the form $1/\infty$ which is 0
  
  f\textsubscript{max,min}(x) ;

\( f_{max,min} \left( x \right)  = max \left( min \left( f(x, \theta) \right)   \exists i~ \forall j \right)   \forall i \) ; Hence Proved
\section{Setup and Results for simulated regression problem:}

As it is seen in the upcoming subsections that the scaling and normalisations differ for different architectures, appropriate rescaling was done to bring all observed mean squared errors to the scale of z-score normalised y.

\subsection{Setup for Simulated Regression Environment}
 Shown in table 4.
%%%%%%%%%%%%%%%%%%%%%% table 4 %%%%%%%%%%%%
\begin{table}[H]
%   \caption{Setup for Simulated Regression Dataset Details}
\caption{General setup and data generation method common to all regressors.}
%   \label{}
  \centering
  \begin{tabular}{ll}
    % \cmidrule(r){1-2}
    \toprule
    \textbf{Simulated Regression Dataset Details} \\
    \midrule
    Generating Equation & sin( 20x+3 ) + 2x + 1 + sin( 50x+2 )  \\
    & + sin(x) + s(noise) \\
    Amount and Type of Noise & Random Normal, scale: 0.1 \\
    X range for generation & 0 to 1 \\
    Dimensions of X & 1 \\
    Dimensions of Y & 1 \\
    % \bottomrule
    \midrule
    % \cmidrule(r){1-2}
    \textbf{Environment Details}\\
    \midrule
    \textbf{Environment Variable} & \textbf{Variables}\\
    \midrule
    Language & Python \\
    Framework & Google Colaboratory [3] \\
    Ram & 12.4GB \\
    Disk Space & 37GB available \\
    GPU used & Tesla T4 \\
    Number of GPUs & 1 \\
    % \bottomrule
    \midrule
    % \cmidrule(r){1-2}
    \textbf{Training Details} \\
    \midrule
    Runs per reported result & 10 \\
    \bottomrule    
  \end{tabular}
\end{table}
%%%%%%%%%%%%%%% table ends %%%%%%%%%%%%%%%%%%%%%%%

\subsection{For Linear log-exp form:}
 Shown in table 5.
%%%%%%%%%%%%%%%%%%%%%% table 5 %%%%%%%%%%%%
\begin{table}[ht]
%   \caption{Setup for Simulated Regression Dataset Details}
\caption{Hyperparameters for linear log-exp form regressors. These are decided using manual search.}
%   \label{}
  \centering
  \begin{tabular}{ll}
    % \cmidrule(r){1-2}
    \toprule
    \textbf{Variable} & \textbf{Value}  \\
    \midrule
    Epochs & 2000 \\
    Learning Rate & 1e-2(best for architecture) \\
    Parameters & 150 \\
    Value of a & 10 \\
    Optimizer & Adam \\
    No. of unions and intersections & Unions=25, Intersections=3 \\
    Y scaling & Uniform linear 0 to 1 (worked best) \\
    X scaling & z-score \\
    Loss &MSE \\
    \bottomrule    
  \end{tabular}
\end{table}
%%%%%%%%%%%%%%% table ends %%%%%%%%%%%%%%%%%%%%%%%

\subsection{For Sinusoidal log-exp form:}
Shown in table 6
%%%%%%%%%%%%%%%%%%%%%% table 6 %%%%%%%%%%%%
\begin{table}[ht]
%   \caption{Setup for Simulated Regression Dataset Details}
\caption{Hyperparameters for sinusoidal log-exp form regressors. These are decided using manual search.}
%   \label{}
  \centering
  \begin{tabular}{ll}
    % \cmidrule(r){1-2}
    \toprule
    \textbf{Variable} & \textbf{Value}  \\
    \midrule
    Epochs & 2000 \\
    Learning Rate & 1e-2(best for architecture) \\
    Parameters & 396 \\
    Value of a & 10 \\
    Optimizer & Adam \\
    No. of unions and intersections & Unions=33, Intersections=3 \\
    Y scaling & Uniform linear 0 to 1 (worked best) \\
    X scaling & z-score \\
    Loss & MSE \\
    \bottomrule    
  \end{tabular}
\end{table}
%%%%%%%%%%%%%%% table ends %%%%%%%%%%%%%%%%%%%%%%%

\subsection{For 1 hidden NN}
Shown in table 7
%%%%%%%%%%%%%%%%%%%%%% table 7 %%%%%%%%%%%%
\begin{table}[ht]
%   \caption{Setup for Simulated Regression Dataset Details}
\caption{Hyperparameters for 1 hidden layer NN regressors. These are decided using manual search.}
%   \label{}
  \centering
  \begin{tabular}{ll}
    % \cmidrule(r){1-2}
    \toprule
    \textbf{Variable} & \textbf{Value}  \\
    \midrule
    Epochs & 2000 \\
    Batch size & 64 \\
    Learning Rate & 1e-2 \\
    Parameters & 220 \\
    Activation function & Leaky Relu \\
    Optimizer & Adam \\
    Structure & [1,73,1] \\
    Y scaling & Z-Score normalization(Best Working) \\
    X scaling & Z-Score normalization(Best Working) \\
    Loss & MSE \\
    \bottomrule    
  \end{tabular}
\end{table}
%%%%%%%%%%%%%%% table ends %%%%%%%%%%%%%%%%%%%%%%%
\subsection{for 3 hidden NN}
Shown in table 8
%%%%%%%%%%%%%%%%%%%%%% table 8 %%%%%%%%%%%%
\begin{table}[ht]
%   \caption{Setup for Simulated Regression Dataset Details}
\caption{Hyperparameters for 3 hidden layer NN regressors. These are decided using manual search.}
%   \label{}
  \centering
  \begin{tabular}{ll}
    % \cmidrule(r){1-2}
    \toprule
    \textbf{Variable} & \textbf{Value}  \\
    \midrule
    Epochs & 2,000 \\
    Batch size & 64 \\
    Learning Rate & 1e-2 \\
    Parameters & 390 \\
    Activation function & Leaky Relu \\
    Optimizer & Adam \\
    Structure & [1,12,15,10,1] \\
    Y scaling & Z-Score normalization(Best Working) \\
    X scaling & Z-Score normalization(Best Working) \\
    \bottomrule    
  \end{tabular}
\end{table}
%%%%%%%%%%%%%%% table ends %%%%%%%%%%%%%%%%%%%%%%%
\section{Setup and Results for simulated classification problem:}
\subsection{Setup for Simulated Classification Problem}
Shown in table 9.
%%%%%%%%%%%%%%%%%%%%%% table 9 %%%%%%%%%%%%
\begin{table}[ht]
%   \caption{Setup for Simulated Regression Dataset Details}
\caption{Setup and model details for simulated classifier. Hyperparameters were found using manual search.}
%   \label{}
  \centering
  \begin{tabular}{ll}
    % \cmidrule(r){1-2}
    \toprule
    \textbf{Simulated Regression Dataset Details} \\
    \midrule
    Generating Equation for class A & X: 5e\textsuperscript{-t}sin(15t)+noise; t goes from 0 to 1 \\
    & Y: 5e\textsuperscript{-t}cos(15t)+noise; t goes from 0 to 1 \\
    Generating Equation for class B & X:  4e\textsuperscript{-1.05t}sin(15t)+noise; t goes from 0 to 1 \\
    & Y: 4e\textsuperscript{-1.05t}cos(15t)+noise; t goes from 0 to 1 \\
    Amount and Type of Noise & Random Normal; scale: 0.05 \\
    Examples in A & 10,000 \\
    Examples in B & 10,000 \\
    Train-Test split & 80\% train; 20\% test \\
    Dimensions of input & 2 \\
    Dimensions of Y & 2 categories \\
    % \bottomrule  
    \midrule
    % \cmidrule(r){1-2}
    \textbf{Environment Details} \\
    \cmidrule(r){1-2}
    \textbf{Environment Variable} & \textbf{Values} \\
    \midrule
    Language & Python \\
    Framework & Google Colaboratory [3] \\
    Number of GPUs & 1 \\
    \bottomrule
    \cmidrule(r){1-2}
    \textbf{Training Details} \\
    \midrule
    Y Normalisation & Z-score normalisation \\
    X Normalisation & Z-score normalisation \\
    Loss & Cross Entropy \\
    Epochs & 10,000 \\
    Parameter ‘a’ for linear log-exp form & 1 \\
    Optimizer & Adam \\
    Learning Rates & 5e-2 \\
    Runs per reported result & 10 \\
    \bottomrule
  \end{tabular}
\end{table}
%%%%%%%%%%%%%%% table ends %%%%%%%%%%%%%%%%%%%%%%%

\subsection{Results:}
Shown in table 10 and Figure 12.
%%%%%%%%%%%%%%%%%%%%%% table 12 %%%%%%%%%%%%
\begin{table}[ht]
%   \caption{Setup for Simulated Regression Dataset Details}
\caption{Variation of accuracy with number of unions and intersections}
%   \label{}
  \centering
  \begin{tabular}{ll}
    % \cmidrule(r){1-2}
    \toprule
    u, (i=64/u) & Accuracy(\%) \\
    \midrule
    1 & 75 \\
    2 & 80 \\
    4 & 100 \\
    8 & 100 \\
    16 & 100 \\
    32 & 99 \\
    64 & 73 \\
    \bottomrule
  \end{tabular}
\end{table}
%%%%%%%%%%%%%%% table ends %%%%%%%%%%%%%%%%%%%%%%%

%%%%%%%%%%%%%%%%%%%% Figure 12 %%%%%%%%%%%%%%%%%%%%
\begin{figure}[H]
\centering
		\includegraphics[width=3.43in,height=2.12in]{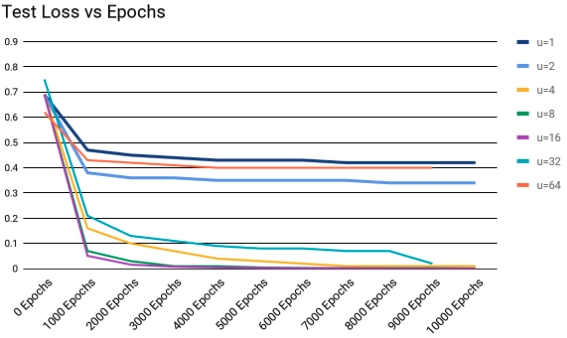}
		\caption{Convergence graphs for various numbers of unions}
\end{figure}
%%%%%%%%%%%%%%%%%%%% Figure Ends here %%%%%%%%%%%%%%%%%%%%

\section{Failures of the max-min form}
The max-min form was trained from scratch on the simulated regression tasks and the results were observed. It was found that the linear components that were already expressed were learnt and few or no new linear components were observed. Some examples are shown in Table 11 but many more runs were done.
%%%%%%%%%%%%%%%%%%%%%% table 11 %%%%%%%%%%%%
\begin{table}[ht]
%   \caption{Setup for Simulated Regression Dataset Details}
\caption{Examples of failure cases for max-min form when trained directly}
%   \label{}
  \centering
  \begin{tabular}{llll}
    % \cmidrule(r){1-4}
    \toprule
     Run no: & Final MSE Error & After Training & Before Training\\
    \midrule
    1 & 0.0179 & \includegraphics[width=1.5in,height=1in]{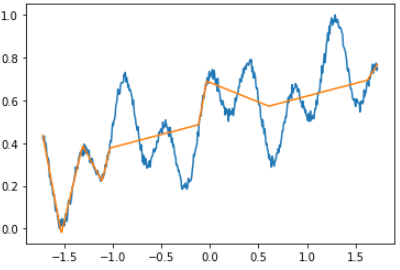} & \includegraphics[width=1.5in,height=1in]{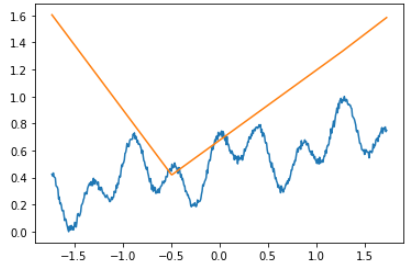} \\
    2 & 0.0236 & \includegraphics[width=1.5in,height=1in]{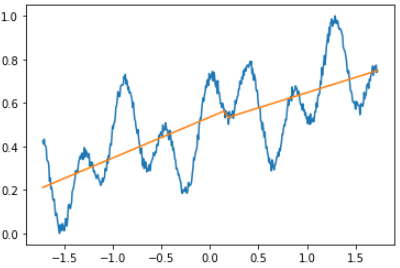} & \includegraphics[width=1.5in,height=1in]{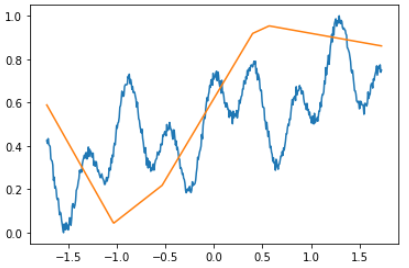} \\
    3 & 0.0065 & \includegraphics[width=1.5in,height=1in]{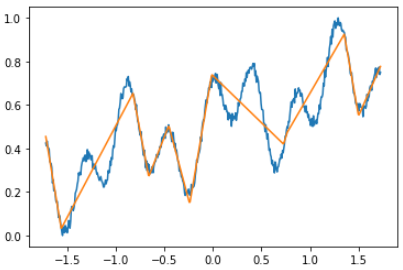} & \includegraphics[width=1.5in,height=1in]{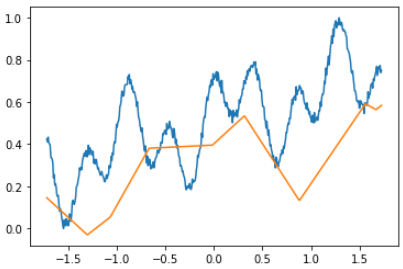} \\
    \bottomrule
  \end{tabular}
\end{table}
%%%%%%%%%%%%%%% table ends %%%%%%%%%%%%%%%%%%%%%%%
\section{Noise analysis of Regressor}
Extra random normal noise was added to the simulated regression data and the accuracy of the model was observed. The accuracy was also compared with noiseless data to verify that the model does not overfit and its proximity with actual generating function. This is shown in Table 12.
%%%%%%%%%%%%%%%%%%%%%% table 12 %%%%%%%%%%%%
\begin{table}[H]
%   \caption{Setup for Simulated Regression Dataset Details}
\caption{Noise Analysis (MSEs are for 0 to 1 linear scaled data)}
%   \label{}
  \centering
  \begin{tabular}{lll}
    % \cmidrule(r){1-3}
    \toprule
     Introduced Noise Scale & MSE Loss with respect & MSE Loss with respect\\
     & to noiseless & to noisy \\
     & curve(x10\textsuperscript{-6}) & curve(x10\textsuperscript{-3}) \\
    \midrule
    0.1 & 8 & 0.2 \\
    0.2 & 12.9 & 0.8 \\
    0.3 & 21 & 1.6 \\
    0.4 & 27.8 & 2.4 \\
    0.5 & 38.5 & 3.4 \\
    0.6 & 44.6 & 4.1 \\
    0.7 & 45.8 & 4.9 \\
    0.8 & 52.2 & 5.9 \\
    0.9 & 63.3 & 6.8 \\
    \bottomrule
  \end{tabular}
\end{table}
%%%%%%%%%%%%%%% table ends %%%%%%%%%%%%%%%%%%%%%%%

%%%%%%%%%%%%%%%%%%%% Figure/Image No: 13 %%%%%%%%%%%%%%%%%%%%
\begin{figure}[ht]
\centering
		\includegraphics[width=5in,height=1.5in]{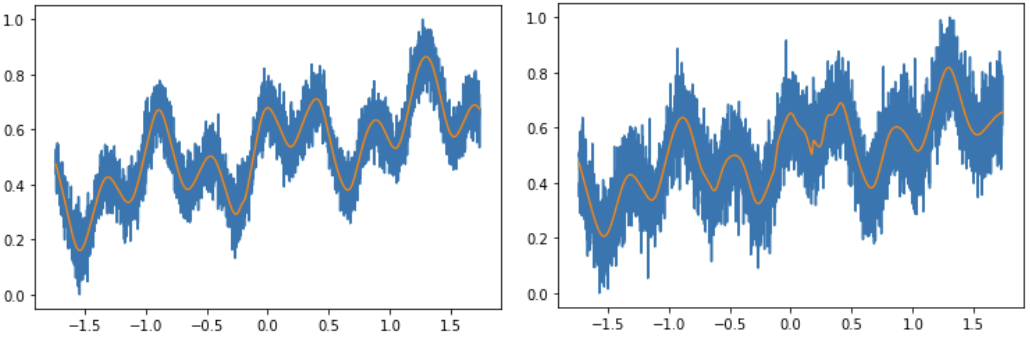}
		\caption{Fit as noise is increased from 0.1 to 0.9 scale. Left: 0.5 noise; Right: 0.9 Noise (Signs and effects of noise are clearly visible)}
\end{figure}
%%%%%%%%%%%%%%%%%%%% Figure Ends here %%%%%%%%%%%%%%%%%%%%

\section{Mixture Piecewise functions}
Mixtures of different components in a polytope can also be fit. This was achieved by defining the model as follows:
$S= (A\textsubscript{1} \cap B\textsubscript{1}) \cup (A\textsubscript{2} \cap B\textsubscript{2}) \cup (A\textsubscript{3} \cap B\textsubscript{3}) \cup ... (A\textsubscript{n} \cap B\textsubscript{n})$
‘n’ was selected to be 10 and set A was selected to be a learnable linear inequality and set B was selected to be a learnable quadratic inequality. This was then expressed in the log-exp form and the resulting equation was fit on the simulated regression problem. The fit can be observed in the Figure 14. It is not necessary that all linear functions are expressed, nor is it necessary for all quadratic inequalities to be expressed. It is not even necessary for any A or any B to be expressed and is decided on the basis of training.
%%%%%%%%%%%%%%%%%%%% Figure 14 %%%%%%%%%%%%%%%%%%%%
\begin{figure}[H]
\centering
		\includegraphics[width=3in,height=1in]{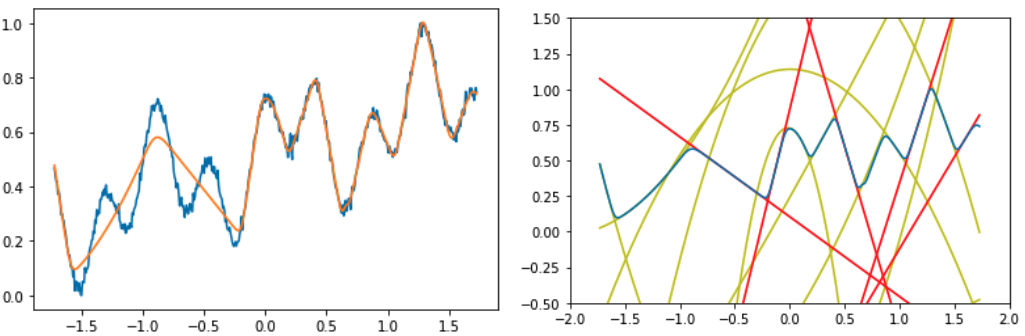}
		\caption{The final fit is not very accurate as it can be seen in this image but includes various components; Left: Blue:Data, Orange:Fit; Right: Blue: Fit; Red: Linear Components; Yellow: Quadratic Components}
\end{figure}
%%%%%%%%%%%%%%%%%%%% Figure Ends here %%%%%%%%%%%%%%%%%%%%

\section{Shared Polytopes}

%%%%%%%%%%%%%%%%%%%% Figure 14 %%%%%%%%%%%%%%%%%%%%
\begin{figure}[H]
\centering
		\includegraphics[width=4in,height=2.5in]{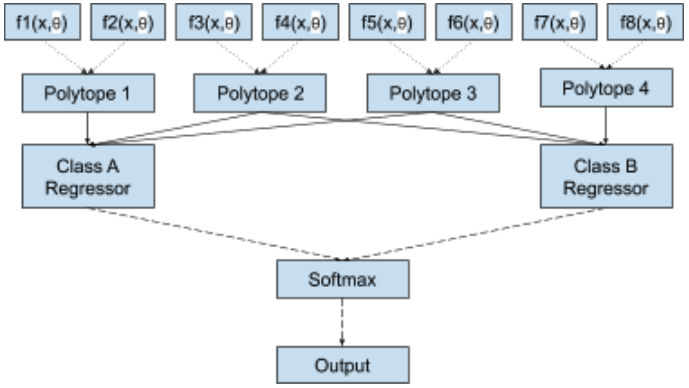}
		\caption{Structure of shared polytope classifier; Dotted arrow: Intersection operation; Complete arrow: Union Operation; Dashed Arrow: Parameter and control flow}
\end{figure}
%%%%%%%%%%%%%%%%%%%% Figure Ends here %%%%%%%%%%%%%%%%%%%%
The diagram above(Figure 15) shows the design of a classifier which has shared polytopes. In this classifier, as stated in the paper, every class has its own regressor. The major difference is that the regressors here can share polytopes among themselves. This significantly reduces the parameters and component functions needed as every regressor will not require separate polytopes. It also increases the capacity of the regressors and is expected to give other advantages as discussed below.

An input now can belong to 3 different states of outputs; Class A, Class B, and equally likely in A and B. Class A is chosen when Polytope 1 gives values greater than other 3 polytopes, class B when Polytope 4 gives values greater than other 3 and the outcome is equally likely in A and B when polytope 2 or 3 give values greater than the rest. Polytopes 1 and 4 are thus called identifying polytopes. A similar example is shown in Figure 16.

This can provide special outcomes in cases where it is difficult to discern the categories or has run out of polytopes to express the given classes as shown in figure below. It may also avoid identifying polytopes of one class affecting identifying polytopes of other classes as softmax increases the probability of one class while reducing that of others thus may affect planes of classes that are not predicted. 

Another major expected advantage is in cases of Hierarchical Classification. There can be a root class which has polytopes in all classes and polytopes common to all classes as identifying polytopes. The root can have subordinate classes(like Vehicle and Food) which have identifying polytopes of their own and common polytopes. Their identifying polytopes include identifying and common polytopes of all child classes. This can then be further extended. Classifications to different depths can be obtained like Class Car would mean an identifying polytopes of Class Car, Vehicle and Root are selected, while class Food would mean identifying polytope of Food and Root are selected but not specific polytopes of classes like Fruit or Vegetable (thus their common polytope is selected). Implementation and testing of this interesting idea is left for future.

Many more advantages in terms of complexity control, and scalability are expected from this design aspect of the classifier architecture and will surely be explored further in the future.
%%%%%%%%%%%%%%%%%%%% Figure 15 %%%%%%%%%%%%%%%%%%%%
\begin{figure}[ht]
\centering
		\includegraphics[width=5in,height=2in]{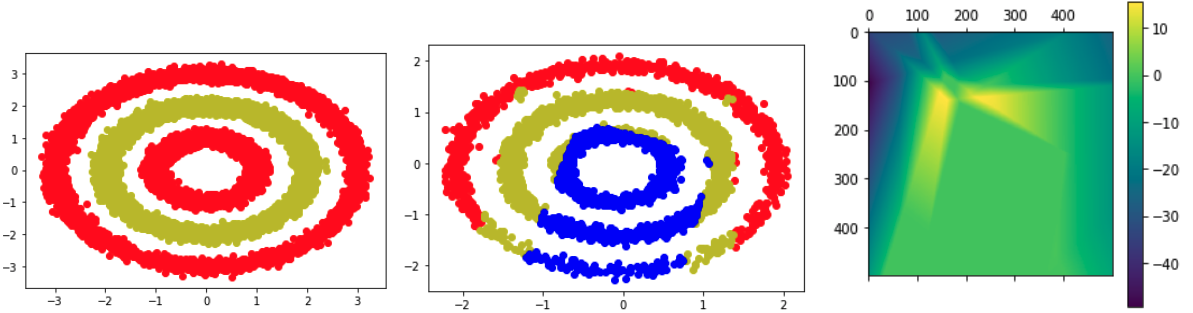}
		\caption{Analysing shared polytope classifier; Left: Actual Dataset, Red: Class A, Green: Class B; Middle: Fit by classifier with 1 polytope per class and 2 shared polytopes, Blue: Either A or B; Right: heatmap for Regressor B - Regressor A }
\end{figure}
%%%%%%%%%%%%%%%%%%%% Figure Ends here %%%%%%%%%%%%%%%%%%%%
\section{Environment and setup for MNIST, FMNIST and CIFAR 10}
Common Setup details are shown in Table 13. All hyperparameters are found by manually trying various combinations.
%%%%%%%%%%%%%%%%%%%%%% table 13 %%%%%%%%%%%%
\begin{table}[H]
%   \caption{Setup for Simulated Regression Dataset Details}
\caption{Common setup and preprocessing for all cases}
%   \label{}
  \centering
  \begin{tabular}{ll}
    % \cmidrule(r){1-2}
    \toprule
     \textbf{Variable} & \textbf{Value} \\
    \midrule
    Preprocessing & Mean = 0.1307, SD=0.3081 \\
    Runs & 10 \\
    \bottomrule
  \end{tabular}
\end{table}
%%%%%%%%%%%%%%% table ends %%%%%%%%%%%%%%%%%%%%%%%
\subsection{MNIST}
\subsubsection{For Linear log-exp form:}
%%%%%%%%%%%%%%%%%%%%%% table 16 %%%%%%%%%%%%
\begin{table}[H]
%   \caption{Setup for Simulated Regression Dataset Details}
\caption{Setup for linear log-exp classifier}
%   \label{}
  \centering
  \begin{tabular}{ll}
    % \cmidrule(r){1-2}
    \toprule
    \textbf{Variable} & \textbf{Value} \\
    \midrule
    Epochs & 20 \\
    Batch Size & 1000 \\
    Learning Rate & 1e-2(best for architecture) \\
    Parameters & 502,400 (even works with 125,600) \\
    Value of a & 0.1 \\
    Optimizer & Adam \\
    No. of Unions and Intersections & Regressors=10, Unions=16, Intersections=4 \\
    Loss & Negative Log Loss \\
    \bottomrule
  \end{tabular}
\end{table}
%%%%%%%%%%%%%%% table ends %%%%%%%%%%%%%%%%%%%%%%%

\subsubsection{For 2 hidden NN}
%%%%%%%%%%%%%%%%%%%%%% table 17 %%%%%%%%%%%%
\begin{table}[H]
%   \caption{Setup for Simulated Regression Dataset Details}
\caption{Setup for 2 hidden layer FCNN classifier}
%   \label{}
  \centering
  \begin{tabular}{ll}
    % \cmidrule(r){1-2}
    \toprule
    \textbf{Variable} & \textbf{Value} \\
    \midrule
    Epochs & 20 \\
    Batch Size & 1000\\
    Learning Rate & 1e-2 \\
    Parameters & 241,575 \\
    Activation function & Relu \\
    Optimizer & Adam \\
    Structure & [784,289,49,10] \\
    Loss & Negative Log Loss \\
    \bottomrule
  \end{tabular}
\end{table}
%%%%%%%%%%%%%%% table ends %%%%%%%%%%%%%%%%%%%%%%%

\subsubsection{For 4 hidden NN}
%%%%%%%%%%%%%%%%%%%%%% table 18 %%%%%%%%%%%%
\begin{table}[H]
%   \caption{Setup for Simulated Regression Dataset Details}
\caption{Setup for 4 hidden layer FCNN classifier}
%   \label{}
  \centering
  \begin{tabular}{ll}
    % \cmidrule(r){1-2}
    \toprule
    \textbf{Variable} & \textbf{Value} \\
    \midrule
    Epochs & 20 \\
    Learning Rate & 1e-2 \\
    Parameters & 311,785 \\
    Activation function & Relu \\
    Optimizer & Adam \\
    Structure & [784,289,189,100,100,10] \\
    Loss & Negative Log Loss \\
    \bottomrule
  \end{tabular}
\end{table}
%%%%%%%%%%%%%%% table ends %%%%%%%%%%%%%%%%%%%%%%%

\subsection{FMNIST}
\subsubsection{For Linear log-exp form:}
%%%%%%%%%%%%%%%%%%%%%% table 19 %%%%%%%%%%%%
\begin{table}[H]
%   \caption{Setup for Simulated Regression Dataset Details}
\caption{Setup for linear log-exp classifier}
%   \label{}
  \centering
  \begin{tabular}{ll}
    % \cmidrule(r){1-2}
    \toprule
    \textbf{Variable} & \textbf{Value} \\
    \midrule
    Epochs & 50 \\
    Batch Size & 1000 \\
    Learning Rate & 1e-2(best for architecture) \\
    Parameters & 502,400 (even works with 125,600)\\
    Value of a & 0.1 \\
    Optimizer & Adam \\
    No. of Unions and Intersections & Regressors=10, Unions=16, Intersections=4 \\
    Loss & Negative Log Loss \\
    \bottomrule
  \end{tabular}
\end{table}
%%%%%%%%%%%%%%% table ends %%%%%%%%%%%%%%%%%%%%%%%

\subsubsection{For 2 hidden NN}
%%%%%%%%%%%%%%%%%%%%%% table 17 %%%%%%%%%%%%
\begin{table}[H]
%   \caption{Setup for Simulated Regression Dataset Details}
\caption{Setup for 2 hidden layer FCNN classifier}
%   \label{}
  \centering
  \begin{tabular}{ll}
    % \cmidrule(r){1-2}
    \toprule
    \textbf{Variable} & \textbf{Value} \\
    \midrule
    Epochs & 50 \\
    Batch Size & 1000\\
    Learning Rate & 1e-2 \\
    Parameters & 241,575 \\
    Activation function & Relu \\
    Optimizer & Adam \\
    Structure & [784,289,49,10] \\
    Loss & Negative Log Loss \\
    \bottomrule
  \end{tabular}
\end{table}
%%%%%%%%%%%%%%% table ends %%%%%%%%%%%%%%%%%%%%%%%

\subsubsection{For 4 hidden NN}

%%%%%%%%%%%%%%%%%%%%%% table 21 %%%%%%%%%%%%
\begin{table}[H]
%   \caption{Setup for Simulated Regression Dataset Details}
\caption{Setup for 4 hidden layer FCNN classifier}
%   \label{}
  \centering
  \begin{tabular}{ll}
    % \cmidrule(r){1-2}
    \toprule
    \textbf{Variable} & \textbf{Value} \\
    \midrule
    Epochs & 50 \\
    Batch Size & 1000\\
    Learning Rate & 1e-2 \\
    Parameters & 293,935 \\
    Activation function & Relu \\
    Optimizer & Adam \\
    Structure & [784,289,189,49,49,10] \\
    Loss & Negative Log Loss \\
    \bottomrule
  \end{tabular}
\end{table}
%%%%%%%%%%%%%%% table ends %%%%%%%%%%%%%%%%%%%%%%%
\subsection{CIFAR 10}

\subsubsection{For Linear log-exp form:}

%%%%%%%%%%%%%%%%%%%%%% table 22 %%%%%%%%%%%%
\begin{table}[H]
%   \caption{Setup for Simulated Regression Dataset Details}
\caption{Setup for linear log-exp classifier}
%   \label{}
  \centering
  \begin{tabular}{ll}
    % \cmidrule(r){1-2}
    \toprule
    \textbf{Variable} & \textbf{Value}  \\
    \midrule
    Epochs & 50 \\
    Batch Size & 1000\\ 
    Learning Rate & 1e-2(best for architecture) \\
    Parameters & 1,114,540 \\
    Value of a & 5e-2 \\
    Optimizer & Adam \\ 
    No. of Unions and Intersections & Regressors=10, Unions=1, Intersections=64,\\ 
    & Linear layer: 3072 to 300 \\
    Loss & Negative Log Loss \\
    \bottomrule    
  \end{tabular}
\end{table}
%%%%%%%%%%%%%%% table ends %%%%%%%%%%%%%%%%%%%%%%%

\subsubsection{For 2 hidden NN}
%%%%%%%%%%%%%%%%%%%%%% table 17 %%%%%%%%%%%%
\begin{table}[H]
%   \caption{Setup for Simulated Regression Dataset Details}
\caption{Setup for 2 hidden layer FCNN classifier}
%   \label{}
  \centering
  \begin{tabular}{ll}
    % \cmidrule(r){1-2}
    \toprule
    \textbf{Variable} & \textbf{Value} \\
    \midrule
    Epochs & 50 \\
    Batch Size & 1000\\
    Learning Rate & 1e-2 \\
    Parameters & 1,707,274 \\
    Activation function & Relu \\
    Optimizer & Adam \\
    Structure & [3072,512,256,10] \\
    Loss & Negative Log Loss \\
    \bottomrule
  \end{tabular}
\end{table}
%%%%%%%%%%%%%%% table ends %%%%%%%%%%%%%%%%%%%%%%%

\subsubsection{For 4 hidden NN}

%%%%%%%%%%%%%%%%%%%%%% table 21 %%%%%%%%%%%%
\begin{table}[H]
%   \caption{Setup for Simulated Regression Dataset Details}
\caption{Setup for 4 hidden layer FCNN classifier}
%   \label{}
  \centering
  \begin{tabular}{ll}
    % \cmidrule(r){1-2}
    \toprule
    \textbf{Variable} & \textbf{Value} \\
    \midrule
    Epochs & 50 \\
    Batch Size & 1000\\
    Learning Rate & 1e-2 \\
    Parameters & 8,444,855 \\
    Activation function & Leaky Relu \\
    Optimizer & Adam \\
    Structure & [3072,2048,1024,49,49,10] \\
    Loss & Negative Log Loss \\
    \bottomrule
  \end{tabular}
\end{table}
%%%%%%%%%%%%%%% table ends %%%%%%%%%%%%%%%%%%%%%%%

\subsection{Important Results and Convergence Graphs:}
%%%%%%%%%%%%%%%%%%%% Figure/Image No: 14 %%%%%%%%%%%%%%%%%%%%
\begin{figure}[H]
\centering
		\includegraphics[width=4in,height=1.5in]{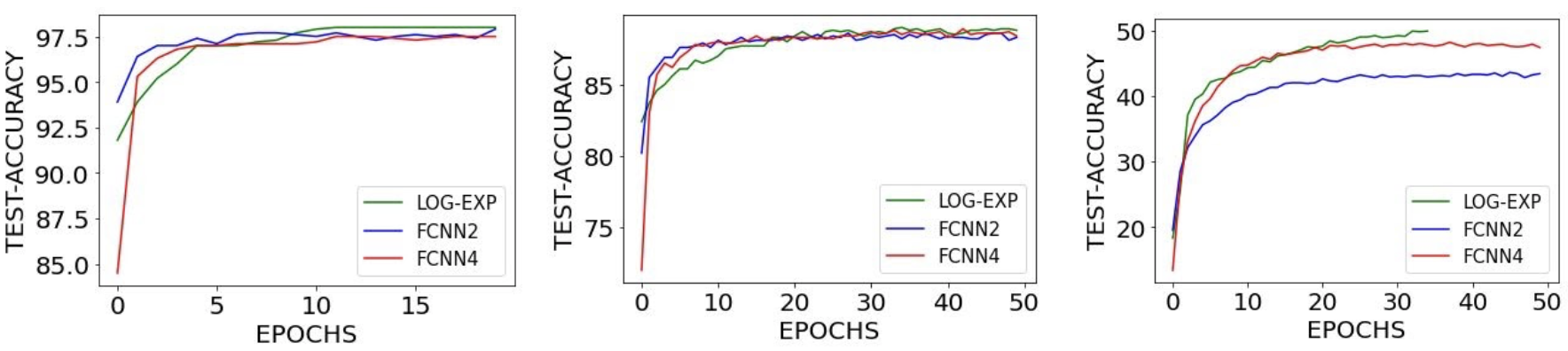}
		\caption{Convergence on Benchmarks: Left:MNIST; Middle:FMNIST; Right: Cifar10}
\end{figure}
%%%%%%%%%%%%%%%%%%%% Figure Ends here %%%%%%%%%%%%%%%%%%%%

%%%%%%%%%%%%%%%%%%%% Figure/Image No: 15 %%%%%%%%%%%%%%%%%%%%
\begin{figure}[ht]
\centering
		\includegraphics[width=3in,height=1.5in]{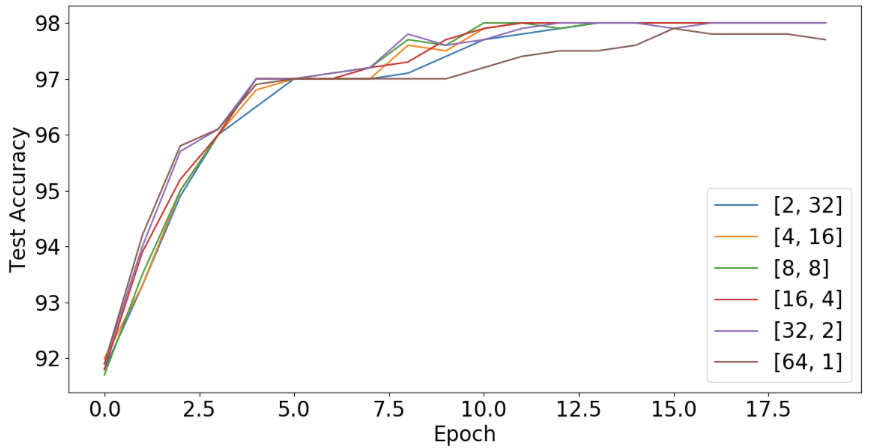}
		\caption{Behaviour of different u and i on MNIST}
\end{figure}
%%%%%%%%%%%%%%%%%%%% Figure Ends here %%%%%%%%%%%%%%%%%%%%

%%%%%%%%%%%%%%%%%%%% Figure/Image No: 16 %%%%%%%%%%%%%%%%%%%%
\begin{figure}[ht]
\centering
		\includegraphics[width=3in,height=1.5in]{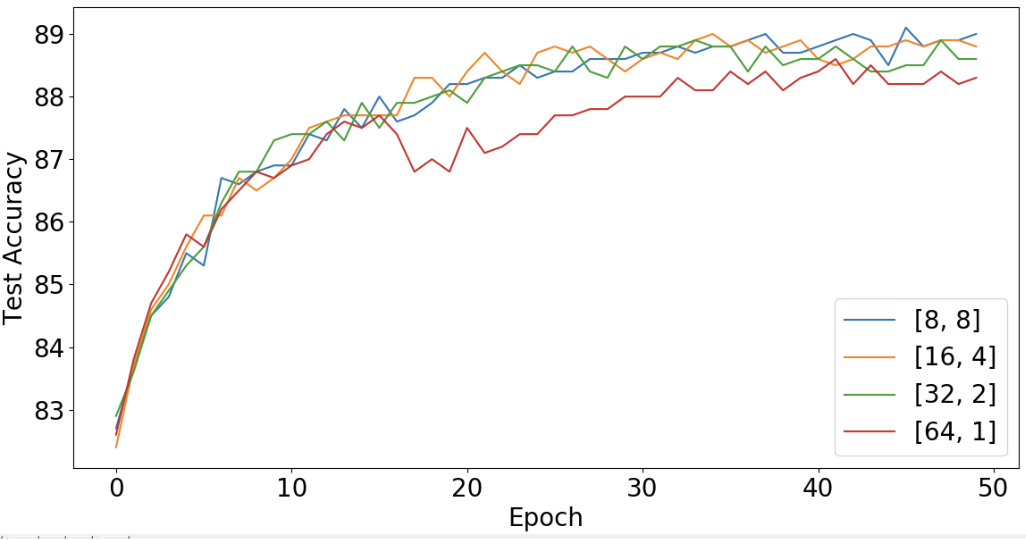}
		\caption{Behaviour of different u and i on FMNIST}
\end{figure}
%%%%%%%%%%%%%%%%%%%% Figure Ends here %%%%%%%%%%%%%%%%%%%%

\section{Major Sinusoidal Components:}
These are some easily interpretable sinusoidal components but all are not so interpretable. This strongly depends on the structure of the component function and also on the value of ‘a’ used. 

%%%%%%%%%%%%%%%%%%%% Figure/Image No: 17 %%%%%%%%%%%%%%%%%%%%
\begin{figure}[H]
\centering
		\includegraphics[width=3in,height=3.5in]{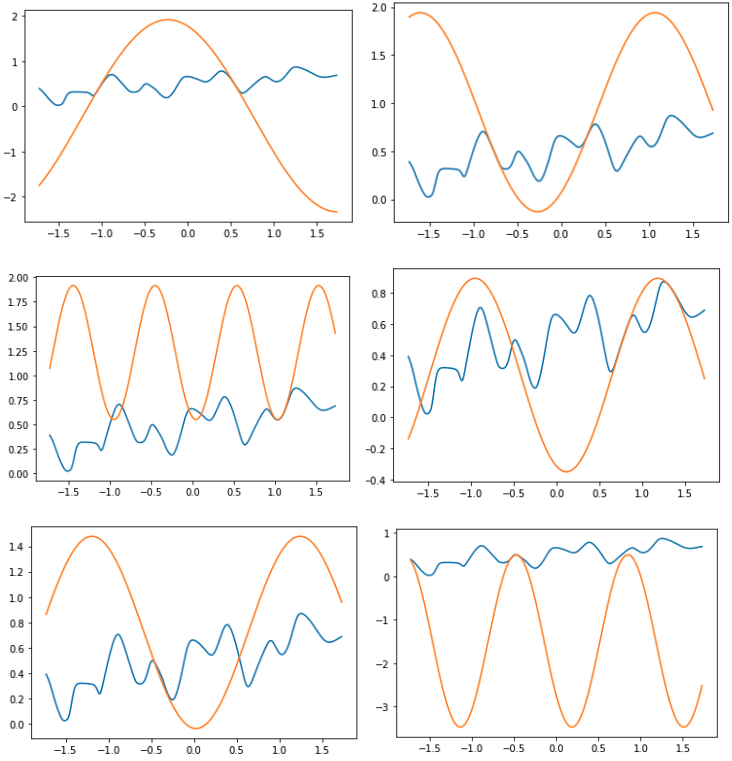}
		\caption{Some easily interpretable sinusoidal components}
\end{figure}
%%%%%%%%%%%%%%%%%%%% Figure Ends here %%%%%%%%%%%%%%%%%%%%

\section{Important Codes}
This section has python code for all important class definitions used in the paper. All implementation details and experimentation code using these classes is in the provided jupyter notebook.

\subsection{Log-exp Regressor:}
\begin{lstlisting}[language=Python]
class LERegressor(nn.Module):
  # defining the model
  # note that a=10 is used for clearly visualising every component
    def __init__(self):
        super(LERegressor, self).__init__()
        self.fc1 = nn.Linear(1, 75) # 1D input
        # 75 linear component functions used 
        # to approximate target function

    def forward(self, x):
        self.x = x
        self.after_fc1 = self.fc1(self.x)
        self.after_fc1_reshaped =
        -1e1*self.after_fc1.view(-1,25,3)
        # reshaping to separate intersecting components
        # and multiplying with a
        # '-' can be absorbed in parameters
        self.expos = torch.exp(self.after_fc1_reshaped)
        # taking exponent
        self.sums = torch.sum(self.expos,axis=2)
        # performing inner summation for intersection
        self.inv_sums = 1/self.sums # taking inverse
        self.total = torch.sum(self.inv_sums,axis=1)
        # performing union operation
        self.logs = torch.log(self.total)/1e1
        # taking log and dividing with a
        return self.logs # using this as the prediction
\end{lstlisting}

\subsection{Log-exp Classifier:}
\begin{lstlisting}[language=Python]
class LEClassifier(nn.Module):
    # defining the model
    def __init__(self):
        super(LEClassifier, self).__init__()
        self.fc1 = nn.Linear(2, 60)
        # 2 input variables and 60 linear component functions

    def forward(self, x):
        self.x = x
        self.after_fc1 = self.fc1(self.x)
        self.after_fc1 = 1*self.after_fc1.view(-1,2,10,3)
        # 2 classes, 10 unions of intersections of 
        # 3 linear components each
        # using a=1
        self.expos = torch.exp(self.after_fc1)
        # '-' sign is assumed to be learnt by parameters
        # taking exponent
        self.sums = torch.sum(self.expos,axis=3)
        # performing sum for intersection
        self.inv_sums = 1/self.sums
        #inverting sums
        self.total = torch.sum(self.inv_sums,axis=2)
        # taking sum for union
        self.logs = torch.log(self.total)/1
        # taking log and dividing by a
        self.prob = F.softmax(self.logs)
        # taking softmax
        return self.prob
\end{lstlisting}
\subsection{Max-min Regressor:}
\begin{lstlisting}[language=Python]
class MMRegressor(nn.Module):
  # defining the model
    def __init__(self):
        super(MMRegressor, self).__init__()
        self.fc1 = nn.Linear(1, 75)
        # 75 component functions used with 1D input
        
    def forward(self, x):
        self.x = x
        self.after_fc1 = self.fc1(self.x) 
        # computing linear components
        self.after_fc1_reshaped = self.after_fc1.view(-1,25,3)
        # first 3 functions are being intersected and then the
        # 25 sets are being unioned
        self.final = torch.max(torch.min(
        self.after_fc1_reshaped, 2)[0], 1)[0] 
        # implementing unions and intersections using max-min
        return self.final
\end{lstlisting}
\subsection{Max-min Classifier:}
\begin{lstlisting}[language=Python]
class MMClassifier(nn.Module):
    def __init__(self):
        super(MMClassifier, self).__init__()
        self.fc1 = nn.Linear(2, 60)
        # 2 input variables and 60 linear component functions

    def forward(self, x):
        self.x = x
        self.after_fc1 = self.fc1(self.x)
        self.after_fc1 = self.after_fc1.view(-1,2,10,3)
        # 2 classes, 10 unions of intersections
        # of 3 linear components each
        self.final = torch.max(torch.min(self.after_fc1,
        axis=3)[0], axis=2)[0]
        self.prob = F.softmax(self.final)
        return self.prob
\end{lstlisting}

\subsection{Sinusoidal Component Regressor:}
\begin{lstlisting}[language=Python]
class SinusoidalRegressor(nn.Module):
    #This defines the structure of the regressor.
    def __init__(self):
        super(SinusoidalRegressor, self).__init__()
        self.fc1 = nn.Linear(1, 99)
        # 99  total component functions
        self.A = Variable(nn.init.normal(torch.empty(99))).cuda()
        self.B = Variable(nn.init.normal(torch.empty(99))).cuda()

    def forward(self, x):
        self.x = x
        self.after_fc1 = self.fc1(self.x)
        # computing a linear function
        self.sine = self.A*torch.sin(self.after_fc1)+self.B
        # making it sinusoidal
        self.sine_reshape = 5*self.sine.view(-1,33,3)
        # every 3 contiguous components are intersected 
        # and then results are unioned
        # multiplying with a=5
        self.expos = torch.exp(self.sine_reshape)
        # taking exponent
        self.sums = torch.sum(self.expos,axis=2)
        # taking sum to perform intersection
        self.inv_sums = 1/self.sums
        # taking inverse
        self.total = torch.sum(self.inv_sums,axis=1)
        # taking sum to perform union
        self.logs = torch.log(self.total)/5
        # taking log and dividing by a
        return self.logs
\end{lstlisting}
\subsection{Polytope sharing in Classifiers:}
\begin{lstlisting}[language=Python]
class PolyShareClassifier(nn.Module):
    def __init__(self):
        super(PolyShareClassifier, self).__init__()
        self.fc1 = nn.Linear(2, 12) 
        # 2 input variables and 12 linear component functions

    def forward(self, x):
        self.x = x
        self.after_fc1 = self.fc1(self.x)
        self.after_fc1 = 5*self.after_fc1.view(-1,4,1,3)
        # 2 classes, 4 unions of intersections of 3 
        # linear components each using a as 5
        
        # creating all polytopes
        self.expos = torch.exp(self.after_fc1)
        self.sums = torch.sum(self.expos,axis=3)
        self.inv_sums = 1/self.sums
        self.total = torch.sum(self.inv_sums,axis=2)

        # creating unions
        self.sh1 = torch.sum(self.total.T[:2], axis=0)
        # union of shared polytopes
        self.a = torch.sum(self.total.T[2:3], axis=0)
        # union of polytopes of class A
        self.b = torch.sum(self.total.T[3:], axis=0)
        # union of polytopes of class B
        self.a += self.sh1
        # union of shared polytopes with a
        self.b += self.sh1
        # union of shared polytopes with b
        
        self.logs = torch.log(
        torch.cat([self.a.unsqueeze(1), self.b.unsqueeze(1)]
        ,dim=1))/5
        self.prob = F.softmax(self.logs)
        return self.prob
\end{lstlisting}

\subsection{Combination of components in Regressor:}
\begin{lstlisting}[language=Python]
class MixedRegressor(nn.Module):
  # defining the model
  # note that a=15 is used for clearly visualising every component
    def __init__(self):
        super(MixedRegressor, self).__init__()
        self.fc1 = nn.Linear(2, 30)
        # 30 quadratic component functions
        self.fc2 = nn.Linear(1, 30)
        # 30 linear component functions

    def forward(self, x):
        self.x = x

        # making quadratic components 
        self.after_fc1 = 
        self.fc1(torch.cat([self.x**2,self.x], dim=1))
        self.after_fc1_reshaped = 15*self.after_fc1
        # first 3 functions are being intersected and then the
        # 10 sets are being unioned
        self.expos1 = torch.exp(self.after_fc1_reshaped)
        
        # making linear components
        self.after_fc2 = self.fc2(self.x)
        self.after_fc2_reshaped = 15*self.after_fc2
        # first 3 functions are being intersected and then the
        # 10 sets are being unioned
        self.expos2 = torch.exp(self.after_fc2_reshaped)

        # intersecting the two to give mixed polytopes
        self.sums = self.expos1+self.expos2

        # taking union of polytopes
        self.inv_sums = 1/self.sums
        self.total = torch.sum(self.inv_sums,axis=1)
        self.logs = torch.log(self.total)/15
        return self.logs
\end{lstlisting}

\section{Using the Mathematical Formulation to build Neural Decision Trees:}
Neural Decision Trees as stated in the paper involve nodes which make soft decisions based on individual features or certain interpretable aspect of the data. This decision once made is then used to move down the tree and the process is repeated until a leaf node is reached. This leaf node includes the final prediction class. A path followed by an input is essentially the conjunction of all decisions made along it and the entire neural decision tree can be seen as the disjunction of all paths involved in it. Thus nodes can be modeled as linear layers with sigmoid activation having one or more features. The two possible paths can be expressed and result of node $f(w,x)$ and its complement $1-f(w,x)$. Thus a simple neural decision tree can be expressed as min() along a given path and max of all the mins. The max-min form of a simple neural decision tree with 3 nodes is:

$P(Class 1) = max(min(A(w1,x),B(w2,x)),min(1-A(w1,x),C(w3,x)))$

$P(Class 2) = max(min(A(w1,x),1-B(w2,x)),min(1-A(w1,x),1-C(w3,x)))$

The same architecture can be extended to more depth and the classifier is thus able to generate more complex decision boundaries. This max-min form model is converted into log-exp form and trained. The results are shown below.

\begin{figure}[ht]
\centering
		\includegraphics[width=5.5in,height=2in]{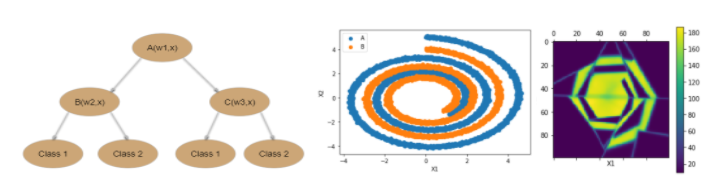}
		\caption{Left: Neural Decision tree expressed above; Middle: Dataset used to test; Right: Decision boundaries of individual nodes}
\end{figure}

\section{Sigmoidal Components in Log-Exp form:}
The log-exp form works very well and slightly better with sigmoidal components than with linear components in certain cases. The structures created in sigmoidal log-exp model are akin to those created in Disjunctive Normal Networks. This can be seen in a small visualization below:

\begin{figure}[H]
\centering
		\includegraphics[width=5.5in,height=2in]{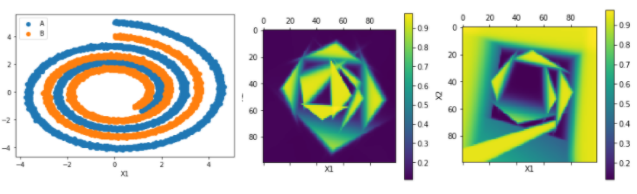}
		\caption{Left: Dataset; Middle: Sigmoidal polytopes of class A; Right: Sigmoidal polytopes of class B}
\end{figure}

\section*{References}
\medskip
\begin{small}

[1] Grover, J. S.\ (2019). Differentiable Set Operations for Algebraic Expressions. arXiv preprint arXiv:1912.12181.

[2] Ovchinnikov, S.\ (2002). Max-min representation of piecewise linear functions. Contributions to Algebra and Geometry, 43(1), 297-302.

[3] https://colab.research.google.com/notebooks/intro.ipynb

\end{small}
\end{document}